\documentclass[11pt]{article}

\usepackage[final]{acl}

\usepackage{times}
\usepackage{latexsym}

\usepackage[T1]{fontenc}
\usepackage[utf8]{inputenc}

\usepackage{microtype}

\usepackage[T1]{fontenc}
\usepackage{inconsolata}

\usepackage{graphicx}
\usepackage{tabularx}
\usepackage{caption}
\usepackage{booktabs}
\usepackage{multirow}
\usepackage{siunitx}
\usepackage{threeparttable}
\usepackage{listings}
\usepackage{caption}
\usepackage{placeins}
\usepackage{needspace}
\usepackage{adjustbox}
\usepackage{makecell}
\usepackage{float}
\usepackage{array}

\usepackage[inline]{enumitem}

\newcolumntype{Y}{>{\centering\arraybackslash}X}

\title{IRON-Meme: A Benchmark for Structured Understanding of Identities and Roles in Online Meme Narratives}

\author{
  \textbf{Gabriella Bollici$^1$}, \textbf{Stefano de Giorgis$^1$}, \textbf{Wouter Van Atteveldt$^2$}, \textbf{Filip Ilievski$^1$} \\
  $^1$Department of Computer Science, Faculty of Science, Vrije Universiteit Amsterdam, \\$^2$Department of Communication Science, Faculty of Social Science, Vrije Universiteit Amsterdam \\
  \texttt{\{a.g.bollicimoreno, s.degiorgis, wouter.van.atteveldt, f.ilievski\}@vu.nl}
}
\begin{document}
\maketitle
\begin{abstract}
As polarization grows on social media, understanding how political memes shape online narratives has become increasingly important. Previous research has relied on either qualitative frameworks that are hard to scale or computational methods that lack theoretical grounding. Both approaches treat narrative categories in isolation and focus exclusively on moral portrayals of identity. To address these limitations, we introduce a framework grounded in social science theories, encompassing three categories: \textbf{social identities} (such as gender, politics, and ethnicity), \textbf{viewpoints} (ingroup and outgroup perspectives), and \textbf{narrative roles} (moral character, intelligence, appearance, and power). Building on this framework, we present IRON-Meme, a human-annotated benchmark comprising 1,110 political memes across substantially more templates than existing benchmarks. Our analysis of IRON-Meme shows that most memes assign derogatory roles and that identities, roles, and templates are systematically interconnected. Evaluation of nine vision-language models indicates that frontier-scale and reasoning models achieve the highest performance, while accuracy on social identities and narrative roles decreases for less frequent labels. Incorporating template background knowledge does not enhance narrative role classification, suggesting that meme interpretation relies on how memes are adapted for specific communicative purposes. We make IRON-Meme and our code publicly available.

\paragraph{Code:} \url{https://github.com/gabriellabollici/ironmeme} 
\end{abstract}

\section{Introduction}

In the digital age, memes have emerged as a widespread and influential form of political communication on social media platforms \citep{AlAfnan}. Memes can be defined as identity-shaping symbols that embody the culture and values of online communities \citep{hennPeekingMemesEvaluating2024}. They communicate group-specific political narratives by drawing on cultural knowledge and shared references, many of them encoded in templates whose conventional uses and communicative functions guide interpretation \citep{joshiContextualizingInternetMemes2024}. In doing so, they reinforce group belonging and collective identity \citep{DeCook, Anderau}. Political memes rarely provide balanced representations of political actors and events, often relying on exaggeration and simplification that produce selective and distorted portrayals \citep{hamidViralMemesExploring2024}. Opposing actors are often depicted as inferior, immoral, unappealing, or incompetent, while members of the favored group are presented as superior, virtuous, attractive, or more knowledgeable \citep{Andreasen, Fiadotava}. These asymmetric depictions are built by the implicit mapping of social identities (e.g. women/men, democrat/conservative) onto narrative roles (e.g. hero, villain), which are constructed through the distinct framing of favored or opposed actors (ingroups or outgroups). Hence, understanding meme narrative construction requires examining the interplay of these three elements: identities, viewpoints and narrative roles, considering the template conventions through which they are conveyed.

Previous studies have introduced theoretical frameworks for analyzing narratives in memes \citep{Saint_Laurent, Godwin, Fiadotava, huntingRolePopularMedia2020} and benchmarks for evaluating the classification of narrative roles \citep{Sharma_2022, barron-cedenoOverviewCLEF2024CheckThat2024, Sharma_Vector, Sharma_WDYM, Sharma_2025}. However, we find four limitations in this work. \textit{First, theoretically grounded computational methods remain scarce and highly fragmented}, as current approaches tend to be either qualitative or scalable, rarely both \citep{Bollici_2026, pehLargeScaleNarrative2026}. \textit{Second, existing research treats social identities, narrative roles, and viewpoints in isolation rather than as an interconnected system}, obscuring the interactions among these narrative categories. \textit{Third, both theoretical and computational approaches have predominantly focused on moral portrayals of identities}, resulting in narrow taxonomies that overlook other vital dimensions of identity construction, such as intellectual ability, physical appearance, and social status. \textit{Fourth, most existing benchmarks do not incorporate template knowledge, and those that do address only a limited set of templates}, leaving open whether such knowledge improves narrative understanding. 

Addressing these gaps requires scalable, theory-grounded frameworks that jointly identify and classify identities, viewpoints, and roles in political memes using broader taxonomies of identity construction, along with benchmarks that cover diverse templates and assess the role of template knowledge in narrative understanding, which is the goal of our paper. We make the following contributions:

% We address these four gaps, %we present IRON-Meme, a benchmark for evaluating whether models can achieve a structured understanding of how political memes construct narratives: who is represented, from whose perspective, and in which role. 
% with our contributions: % are as follows:

\begin{enumerate}
\item \textbf{A unified compositional framework} that jointly structures meme understanding across three narrative categories: a set of social identities (e.g., gender, political affiliation, ideological orientation); a definition of viewpoint based on ingroup/outgroup positioning relative to the author or diffusor; and a narrative role taxonomy that extends beyond moral portrayals to four dimensions: moral, intellectual, physical, and power.
\item \textbf{A novel human-annotated benchmark, IRON-Meme (Identities and Roles in Online Meme Narratives)}, comprising 1,110 template-based memes across 348 unique templates, a much wider template coverage than prior benchmarks. To our knowledge, IRON-Meme is the first dataset to systematically operationalize these narrative categories across diverse political themes and meme templates, thus directly bridging %establishing a direct bridge between 
theoretical conceptualization and empirical evaluation.
\item \textbf{An empirical evaluation of state-of-the-art vision-language models (VLMs)} on IRON-Meme in zero-shot % and in-context learning (ICL) 
settings, assessing the ability of VLMs to scale theory-grounded narrative analysis and the impact of %, including a targeted investigation into whether 
explicit template background knowledge on VLMs' contextual interpretation of narrative roles.

\end{enumerate}

\section{Related work}

\paragraph{Conceptual Frameworks} Recent scholarship has increasingly applied theoretical frameworks to the study of narrative elements in internet memes. The field has moved from Dawkins' view of memes as discrete cultural units to Shifman’s conceptualization of memes as collective sociocultural practices \citep{dawkinsSelfishGene2006, Shifman_2013, galItGetsBetter2016}. %Within this literature, 
Researchers have explored gendered representations and feminist counter-narratives \citep{ArtMemesFeminist, ballebyDigitalIntimacyHow2026}, national and ideologically driven identity formation \citep{mcswineySharingHateMemes2021, Mozolevska}, and perspectives on the climate crisis \citep{ johannFridaysFutureMondays2023, litherlandMemeingWorldBurns2026}. A more specialized strand of research conceptualizes identities as characters occupying certain roles within broader social and political narratives. These approaches, applied to contexts such as COVID-19 \citep{Saint_Laurent}, conspiracy theories \citep{Godwin}, the Russian-Ukrainian war \citep{Fiadotava}, and the 2016 US elections \citep{huntingRolePopularMedia2020}, have primarily examined moral roles, such as hero, villain, and victim, thus limiting the analytical depth and applicability of these frameworks for capturing broader narrative complexities.

\begin{figure*}[t]
\centering
\makebox[\textwidth][c]{\includegraphics[width=0.95\textwidth]{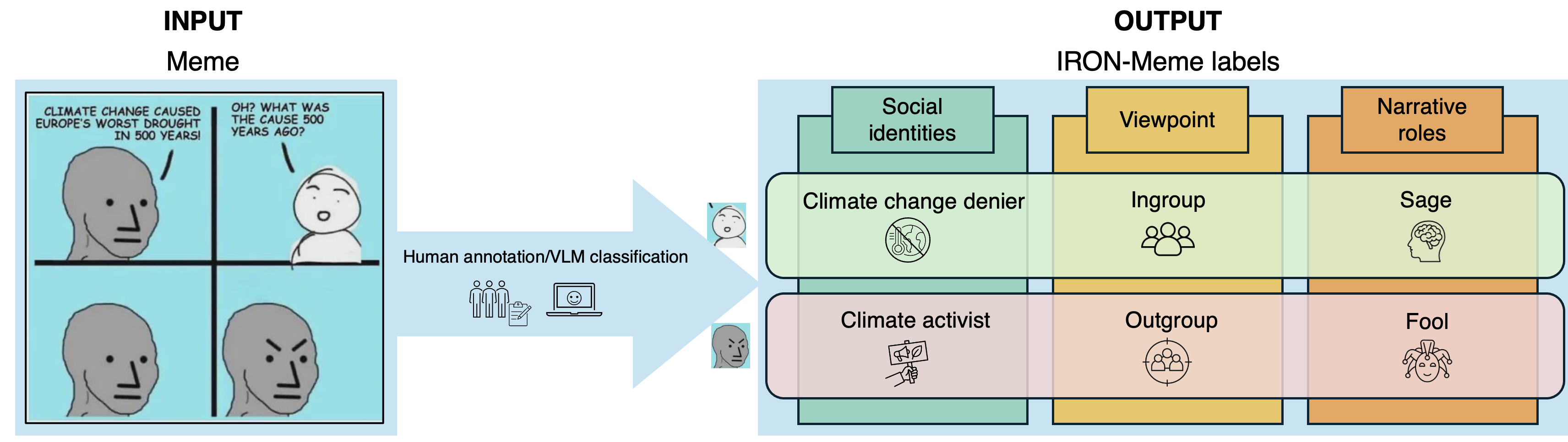}}
\caption{\textbf{Framework overview.} An input meme is annotated by a human or VLM to produce a structured mapping of characters across the three dimensions of the IRON-Meme label schema: social identity, viewpoint, and narrative role. In this example, the white character is assigned the identity \textit{climate change denier} with an \textit{ingroup} viewpoint and \textit{sage} role, while the grey character is classified as a \textit{climate activist} with an \textit{outgroup} viewpoint and \textit{fool} role.}% Together, these categories form the IRON-Meme label schema.}
\label{fig:Framework}
\end{figure*}

\paragraph{Benchmarks} Interest has grown in developing benchmarks to evaluate the classification of social identities and narrative roles in memes. Existing benchmarks vary considerably in their definition of social identity, ranging from broad demographic and social categories to open-ended identification of named or depicted entities \citep{BanglaAbuse, pramanickMOMENTAMultimodalFramework2021, sharmaDISARMDetectingVictims2022}. For narrative role classification, the most prominent benchmarks emerged from the CONSTRAINT 2022 and CheckThat! 2024 shared tasks, which adopt the same moral taxonomy (hero, villain, victim) as the theoretical frameworks \citep{Sharma_2022, barron-cedenoOverviewCLEF2024CheckThat2024, Sharma_2025}. More recently, viewpoint identification has been explored through binary opposing perspectives generated with VLMs \citep{hou_viewpoint}, but to our knowledge, no benchmark has yet been developed for this task. Although these benchmarks establish standardized tasks and evaluation protocols, operationalization of narrative elements is fragmented, and no unified, structured framework yet integrates them. Similarly, existing methods are isolated and task-specific: identity classifiers rely mainly on attention-based multimodal fusion \citep{Ahsan_MIMOSA, hossainDecipheringHateIdentifying2024}, while narrative role classifiers draw on contextual embeddings \citep{Toh}, external knowledge bases \citep{Sharma_Vector, Zhou_CONSTRAINT}, and cross-modal inference \citep{Kun_CONSTRAINT, Singh_CONSTRAINT, Sharma_WDYM}. Finally, most benchmarks do not formalize meme template knowledge at all, and those that do cover only a limited number of templates (e.g., 39 in \cite{Toh}), limiting their ability to assess models' use of templates to interpret political memes.

\section{Framework}

We now detail our framework's elements, which are illustrated in Figure \ref{fig:Framework}.

\noindent \textbf{Social identities} form the foundational level of our framework (Figure~\ref{fig:Framework}): the other two narrative categories, viewpoints and narrative roles, are grounded in, and interpreted through, the identities they represent. Social identities are social categories constructed through discursive practices in which groups are defined by their differences from others, leading to positive or negative differential treatment \citep{wigginsDiscursivePowerMemes2019}. They are based on the definition of shared norms, symbolic references, group representatives, and the distinction between “us” and “them” \citep{Guenther}. Understanding the concept of social identity and this polarized structure is essential for meme analysis \citep{wigginsDiscursivePowerMemes2019}. Identity-defining attributes include gender and gender orientation, ideological orientation, national and ethnic origin, economic status, religious affiliation, and others. 

% viewpoint
\noindent \textbf{Viewpoints} capture the perspective associated with each identity. They are grounded in a central distinction of social identity theory: that between ingroups and outgroups \citep{brownSocialIdentityTheory2000}. Ingroup identities (“us”) are often defined in contrast to outgroup identities (“them”), through the association of traits or characteristics. From the perspective of meme creators, ingroups represent those who belong to their own social category \citep{Guenther}, while outgroups function as antagonistic figures whose voices are systematically excluded, mocked, or targeted \citep{prochazkaCohesiveAspectsHumor2016, milnerFCJ156HackingSocial}. %This contrast can improve individuals' self-image by portraying others as inferior, which may lead to a sense of enjoyment \citep{Immonen, Kidd}.
Accordingly, the viewpoint category distinguishes whether a meme identity represents the perspective of the author or diffusor, identified as the ingroup, or the targeted perspective, identified as the outgroup. 

\noindent \textbf{Narrative roles} capture how identities are portrayed within the narrative. They draw on a crucial insight from character theory: the concept of character can extend beyond individuals, perceived through anthropomorphic projections onto groups, organizations, nations, and similar entities, which are personified and treated as individuals \citep{Bergstrand}. Building on this, we conceptualize social identities in memes as characters that can be assigned narrative roles, increasing memes' potential to exert political influence. Although most existing studies base their narrative role taxonomy on the Drama Triangle by \citet{Karpman}, we argue that this framework does not sufficiently capture portrayals within social media environments, as it focuses exclusively on the moral dimension. For this reason, we expand this framework by introducing a new model based on four dimensions: moral, intellectual, physical, and power (Table \ref{tab:narrative_roles}), drawing upon recurring themes identified in research on online media narratives as well as empirical observations of internet memes:

\begin{table}[t]
\centering
\small\selectfont
\renewcommand{\arraystretch}{1.15}

\caption{\textbf{Narrative role labels in our framework.} This set is based on two sets of binary oppositions (positive/negative and superior/inferior), and four evaluative dimensions (moral, intellectual, physical and power).}
\label{tab:narrative_roles}

\resizebox{\columnwidth}{!}{%
\begin{tabular}{lcccc}
\toprule
\textbf{Evaluative dimension} & \textbf{Moral} & \textbf{Intellectual} & \textbf{Physical} & \textbf{Power} \\
\midrule
\textbf{Positive / Superior} & Hero & Sage & Charmer & Winner \\
\textbf{Negative / Inferior} & Villain & Fool & Monster & Loser \\
\textbf{Positive / Inferior} & Victim &  &  &  \\
\bottomrule
\end{tabular}%
}
\end{table}

\begin{enumerate}
    \item The \textbf{moral dimension} encompasses the hero, the villain, and the victim. \textit{Heroes} are depicted as strong, well-intentioned individuals who protect others and restore order, while \textit{villains} evoke fear and hatred and are usually expected to be punished \citep{Bergstrand, Fiadotava}. \textit{Victims} are portrayed as innocent,  virtuous characters who need protection or rescue \citep{Skrynnikova}. 
    
    \item The \textbf{intellectual dimension} contrasts \textit{the sage}, who is wise, intellectually superior, and self-aware, with \textit{the fool}, who is characterized as irrational, ignorant, unskilled, or prone to poor judgment \citep{Nixon, Godwin, Paz, Mahmoud}. 
    
    \item The roles in the \textbf{physical dimension} are \textit{the charmer}, whose attractiveness aligns with prevailing beauty standards and signals desirability, and \textit{the monster}, whose appearance is depicted as ugly, grotesque, or monstrous \citep{Polli, Moody_Ramirez, Andreasen}.

    \item The \textbf{power dimension} differentiates \textit{winners}, associated with socioeconomic success, education, stability, career achievement, social competence, and fulfilling relationships, from \textit{losers}, who are defined by the absence or opposite of these attributes \citep{Shoebridge, Immonen}. 
\end{enumerate}

Figure~\ref{fig:Framework} presents a meme contrasting two identities based on their perspectives on climate change: the denier, framed as the ingroup the audience is invited to align with, and the activist, targeted as the outgroup whose perspective is dismissed. The denier assumes the role of the sage, using historical questioning as seemingly logical counter-evidence, while the activist is cast as the fool (via the uncritical NPC template), unable to answer and reacting with frustration. However, this is a simplified case: in practice, the same actor may adopt different roles across memes or even within a single meme \citep{Saint_Laurent}, and positive or negative roles may be assigned to ingroups or outgroups (e.g., self-mockery \citep{Immonen}).

\section{IRON-Meme}

This section outlines the construction of the IRON-Meme benchmark, detailing the process from data collection through template-based filtering.

\subsection{Dataset construction}

\begin{figure}[t]
  \centering
  \includegraphics[width=0.7\columnwidth]{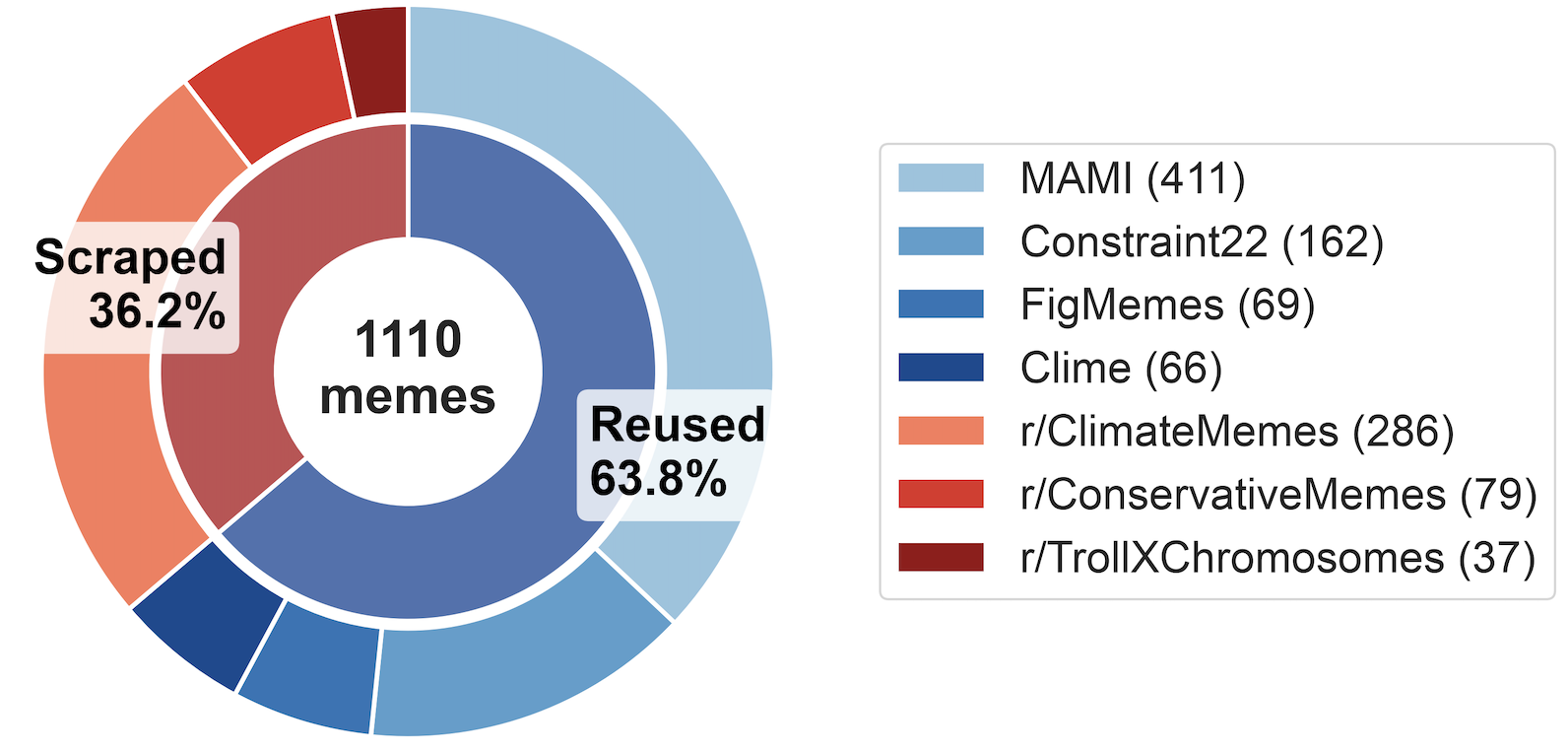}
  \caption{\textbf{Final origin distribution of the IRON-Meme benchmark dataset.} The inner ring illustrates the overall distribution between reused benchmarks (blue) and newly scraped data (red). The outer ring details the individual datasets and subreddits contributing to each category. The legend indicates the number of memes contributed by each source.} 
  \label{fig:dataset_origin}
\end{figure}

\paragraph{Data collection.} The source datasets for IRON-Meme, whose distribution is outlined in Figure \ref{fig:dataset_origin}, were chosen for their diverse domain coverage. 

We reuse four existing datasets introduced in prior work. \textbf{HVVMemes} \cite{pramanickMOMENTAMultimodalFramework2021} provides 3,202 memes from the US politics subset (\textit{Harm-P}), collected from Google Image Search, Reddit, Facebook, and Instagram.
% \item 
\textbf{MAMI} \cite{fersiniSemEval2022Task52022} contributes 11,000 memes focused on anti-women and antifeminist content, collected from 9GAG, Imgur, Reddit, and Twitter.
% \item 
\textbf{CliME} \cite{borahCliMEEvaluatingMultimodal2025} covers 2,579 multimodal posts addressing climate discourse, ranging from humorous commentary to skepticism, sourced from Twitter and climate-focused subreddits such as \texttt{r/climatememes}.
%\item 
\textbf{FigMemes} \cite{Liu_figmemes} provides 5,141 political memes sourced from 4chan's \texttt{/pol/} board, addressing topics such as racial minorities, elections, and LGBTQ+ issues.
% \end{itemize*}

To incorporate more recent memes and ensure that our benchmark reflects up-to-date topics and formats, we collected \textbf{Reddit} data via the PullPush API, retrieving the most recent image submissions available in April 2026. To align with the domains represented in the reused benchmarks, we selected content from r/climatememes, r/conservativememes, and r/trollxchromosomes, focusing on climate discussions, conservative US political perspectives, and feminism, respectively. We collected 5000 memes from each subreddit.  We included feminist content to counterbalance the misogynistic perspectives frequently represented in MAMI and FigMemes.

\paragraph{Template-based filtering.} We retain only memes that match established templates, reducing ambiguity in characterizing their communicative functions. Specifically, we scraped templates from \textit{Know Your Meme} %\footnote{} 
and \textit{Imgflip}.\footnote{www.knowyourmeme.com; www.imgflip.com} We embedded both memes and templates using a DINO ViT-B/16 model, retrieved candidate matches via cosine similarity in FAISS \citep{douzeFaissLibrary2025}, and selected the top template based on a similarity threshold. Remaining unmatched memes were grouped using dHash-based near-duplicate clustering. After manual filtering, the dataset comprised 1,410 memes. More details are available in Appendix \ref{sec:template_based_filtering}.

\subsection{Human annotation and aggregation}
\label{sec:annotation}

\paragraph{Annotation protocol} All memes were independently labeled over a three-month period by three annotators: two female social science master’s students and one male political science bachelor’s student, all aged 24. Annotators used a custom interface (Appendix \ref{sec:annotation_interface}) and a comprehensive dynamic codebook containing category definitions, the analytical framework, multiple annotated examples, and detailed instructions for interface usage and annotation procedures. This process followed the open coding procedure outlined by \citet{sarahj.tracyPhroneticIterativeApproach2018}. Each meme was annotated by all annotators using a hierarchical scheme: they first assigned one or more social identities per meme, then designated a viewpoint, and one or more narrative roles (Table \ref{tab:narrative_roles}) separately for each identified identity. To establish a standardized social identity taxonomy, we conducted an initial dry-run phase in which we iteratively normalized, merged, and expanded freely recorded labels as new categories emerged. This taxonomy stabilized by week 6, when no substantively new categories emerged, producing the final set of 38 social and political identity labels (Appendix \ref{sec:social_identity_labels}). Figure \ref{fig:jaccard_agreement_comparison} visualizes agreement progress across annotation stages.

\paragraph{Derivation of ground-truth labels.} 

Ground-truth labels were derived by retaining annotations agreed on by at least two raters. Of 1,410 annotated memes, 1,112 reached consensus, an overall consensus rate of 78.87\% (30.85\% perfect agreement, 48.01\% majority consensus); after excluding two memes, 1,110 were retained for final analysis. Inter-annotator agreement (IAA) varied markedly across subtasks: viewpoint classification achieved the highest consensus (grand Jaccard index: 0.794), followed by social identities (0.660) and narrative roles (0.532). The lower agreement for narrative roles likely stems from subjective interpretation and implicit group assumptions rather than grounded visual/textual evidence. Detailed agreement statistics, including Fleiss' $\kappa$ and Cohen's $\kappa$, are provided in Appendix~\ref{sec:agreement_metrics}, and examples of ground truth derivation are presented in Table \ref{tab:annotated_examples}.

\subsection{Benchmark statistics and analysis}

\begin{figure}[t]
\centering
\includegraphics[width=0.8\linewidth]{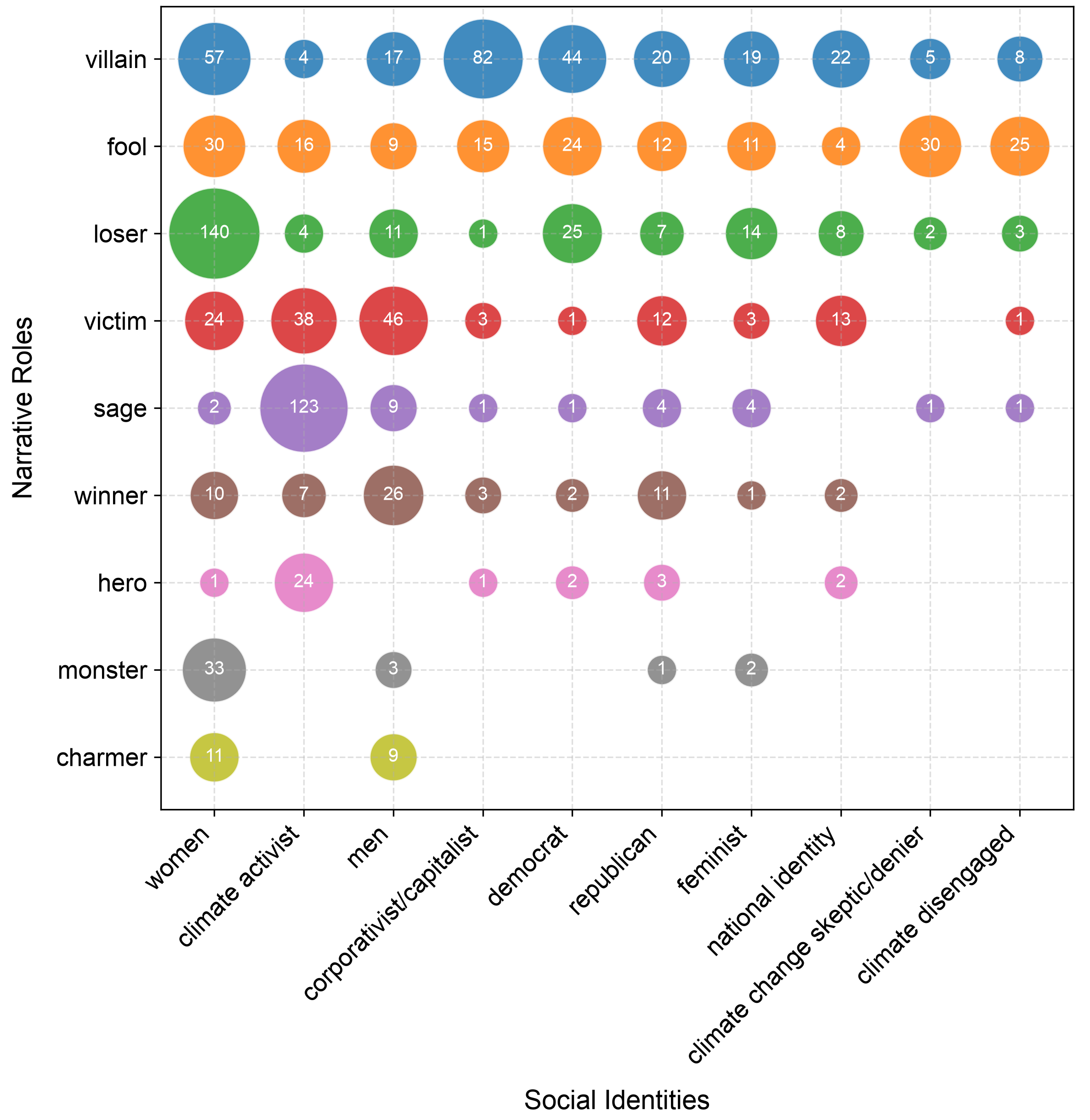}
\caption{\textbf{Identity–role correlation for the 10 most frequent social identities}. The chart displays the frequency of social identity–narrative role combinations, with circle size and numerical values indicating frequency counts. Narrative roles (y-axis) are shown in descending order. %from most to least prominent
Colors correspond to distinct roles.}
\label{fig:role_identities}
\end{figure}

Figure \ref{fig:role_identities} displays the distribution of narrative roles among the ten most frequent social identities. The data show a strong concentration of villain, loser, and fool, indicating that \textit{derogatory roles are particularly prominent}. Notably, 58.92\% of the memes are labeled with negative roles. Looking at the distribution within each identity's assigned roles, corporativist and capitalist identities are mainly depicted as villains (77.4\%), climate skeptics or deniers (78.9\%), and climate-disengaged groups (65.8\%) as fools, and women as losers (45.5\%, vs.\ 25.3\% for democrats and 25.9\% for feminists) and monsters (10.7\%). Conversely, men are mostly represented as victims (35.4\%) and winners (20.0\%), while climate activists are framed as sages (56.9\%). These patterns heavily reflect source dataset sampling rather than general meme culture: 89\% of women annotations and 85\% of feminist annotations stem from MAMI (a misogyny-focused dataset where feminists are 91\% derogatory and 136 of 140 women--loser pairs originate), whereas in female-authored communities (\texttt{r/TrollXChromosomes}), women appear mostly as victims (79\%) and never as losers, with feminists portrayed solely as ingroup sages. Similarly, climate annotations are drawn almost exclusively from pro-climate action sources (\texttt{r/ClimateMemes} and CliME), which frame activists as ingroup sages and skeptics or corporate actors as fools and villains. Section \ref{sec:limitations} details these sampling biases, and Appendix \ref{sec:benchmark_analysis} provides further statistics.

Overall, the distribution indicates that social identities are not represented uniformly across narrative roles; instead, \textit{particular identities are systematically associated with specific evaluations}. Analysis of template usage by category reveals associations suggesting that certain meme formats are repeatedly employed to fulfill specific narrative functions. When examining the distribution of narrative-role labels across meme templates, we observe that the villain role is most frequently associated with \textit{Drakeposting} (16 instances), and  \textit{Distracted Boyfriend} (13), \textit{This Is Fine} with the fool (15), and \textit{My Parents Are Dead / Batman Slapping Robin} with the sage (12). Figure \ref{fig:template_roles} shows the composition of the ten most frequent templates by narrative role. Examining template, narrative role, and social identity jointly reveals more specific recurring configurations. Two templates most frequently represent both men as ingroup victims and women outgroup villains: \textit{Socially Awesome Awkward Penguin} and \textit{Confession Bear}. In the template \textit{My Parents Are Dead / Batman Slapping Robin}, climate activists are frequently constructed as ingroup sages (10), reinforcing the pattern of representing climate activists as knowledgeable or authoritative.

\begin{table*}[t]
\centering
\footnotesize
\renewcommand{\arraystretch}{0.85}
\caption{Zero-shot performance of VLMs across categories (macro metrics) compared against a stratified random baseline. For each meme and each label, the baseline assigns a prediction probability equal to the prevalence of that label in the ground truth. Full model names are listed in Table~\ref{tab:vlm_details}.}
\label{tab:macro_metrics}
\begin{tabular}{l ccc c ccc c ccc}
\toprule
& \multicolumn{3}{c}{\textbf{Social Identities}} & & \multicolumn{3}{c}{\textbf{Viewpoints}} & & \multicolumn{3}{c}{\textbf{Narrative Roles}} \\
\cmidrule{2-4} \cmidrule{6-8} \cmidrule{10-12}
\textbf{Model} & \textbf{Precision} & \textbf{Recall} & \textbf{F1} & & \textbf{Precision} & \textbf{Recall} & \textbf{F1} & & \textbf{Precision} & \textbf{Recall} & \textbf{F1} \\
\midrule
\textit{Random (stratified)} & 11.45 & 11.46 & 11.45 & & 64.94 & 64.96 & 64.94 & & 21.09 & 21.10 & 21.09 \\
\midrule
DeepSeek-VL2 & 11.25 & 39.72 & 13.07 & & 67.18 & 90.89 & 77.10 & & 47.00 & 41.90 & 34.44 \\
Gemini-3.6   & 45.58 & 73.55 & 51.99 & & \bfseries 93.55 & \bfseries 92.86 & \bfseries 93.20 & & \bfseries 59.82 & 75.75 & \bfseries 64.14 \\
GLM-4.1V     & 45.28 & 42.81 & 38.60 & & 81.45 & 77.93 & 79.61 & & 41.20 & 60.49 & 46.55 \\
Kimi-K3      & 44.99 & \bfseries 77.60 & \bfseries 52.78 & & 92.60 & 92.62 & 92.61 & & 48.11 & \bfseries 81.01 & 57.88 \\
Molmo-7B     & 4.16  & 27.99 & 5.80  & & 60.01 & 91.25 & 71.08 & & 17.31 & 15.49 & 11.70 \\
Pixtral-12B  & 36.83 & 47.33 & 35.58 & & 77.29 & 78.72 & 77.98 & & 41.65 & 46.81 & 40.60 \\
Qwen2.5-VL   & 40.58 & 37.88 & 26.41 & & 79.00 & 69.18 & 72.78 & & 34.86 & 32.86 & 23.52 \\
Qwen3-VL-I   & 39.66 & 50.46 & 36.35 & & 76.11 & 80.56 & 77.05 & & 41.30 & 55.53 & 41.74 \\
Qwen3-VL-T   & \bfseries 47.99 & 47.33 & 43.42 & & 80.51 & 75.63 & 77.93 & & 41.54 & 58.57 & 45.51 \\
\bottomrule
\end{tabular}
\end{table*}
\section{Experimental setup}
\paragraph{Evaluation protocol.}
We evaluate nine VLMs following the same hierarchical structure as the human annotation (Section~\ref{sec:annotation}): models first identify the discussion domain and social identities, then assign a viewpoint to each identity, and finally infer identity's narrative roles by selecting an evaluative dimension (moral, intellectual, physical, or power) and polarity (positive or negative) before assigning the role label. All models receive identical inputs, instructions, label inventories, and output constraints. To isolate performance at each stage, prompts are conditioned on upstream ground-truth labels. Model outputs are parsed and normalized to each task's label inventory; unparseable responses and labels outside the inventory are counted as incorrect. Predicted viewpoints and roles are counted as correct only when linked to the correct identity. Parsing failures and discarded labels per model are detailed in Appendix~\ref{sec:parsing}, and prompt templates are listed in Appendix~\ref{sec:prompts}.

Furthermore, we conduct an ablation study using the best-performing model. Considering that meme templates have conventional uses and communicative functions, we hypothesize that knowledge of these conventions helps interpret the narrative roles implicit in a meme. In particular, template knowledge can support a model's inference about the attitudes identities hold toward one another, thus facilitating the classification of narrative roles (Figure \ref{fig:template_roles}). To test this hypothesis, we introduce template knowledge as an information intervention and test whether incorporating conventional meanings and uses improves performance compared to meme-only inputs. We sourced template information for each meme from \textit{Know Your Meme}.

\paragraph{Metrics.} We report precision, recall, and F1-score, each micro- and macro-averaged. Micro-averaging reflects performance across all instances, while macro-averaging weights each class equally, making it more sensitive to performance on less frequent classes, such as the monster/charmer roles or the authoritarian and politically disillusioned identity labels. We use macro-F1 as our primary metric. See Appendix~\ref{sec:metrics} for details and formulas.

\paragraph{VLM selection.}
We conducted a zero-shot ICL evaluation using nine state-of-the-art VLMs that represent a range of \textit{model families, architectures, and reasoning capabilities}. We also include geographically diverse models, accounting for how differing alignment procedures, training data, and development contexts affect performance in socially and narratively complex visual tasks.
A complete overview of the models and their respective specifications is given in Appendix ~\ref{sec:experimental_results}.

\section{Experimental results}
% \subsection{VLM comparison}

% \paragraph{Experimental questions.}
We address four questions:
$(\textit{RQ}_1)$ How does performance differ across models and tasks?
$(\textit{RQ}_2)$ What are the underlying causes of model performance?
$(\textit{RQ}_3)$ How does the frequency of specific meme templates, social identities, and narrative roles within the dataset influence model performance?
$(\textit{RQ}_4)$ How does explicit knowledge about a meme template's conventional meaning and usage affect a VLM's inference of narrative roles?

\paragraph{$\textit{RQ}_1$: Variations in performance across models and tasks}
Performance varies substantially across both models and tasks (Table~\ref{tab:macro_metrics}). Viewpoint classification is the easiest task, with all VLMs exceeding 70\% F1, reflecting its binary label space. Social identity and narrative role classification are considerably more challenging, with scores ranging from 5.80 to 52.78 and from 11.70 to 64.14 F1, respectively. Scores on these two tasks are highly correlated (Spearman's $\rho = 0.95$), with DeepSeek-VL2 as the main exception (13.07 vs.\ 34.44). 

Two models clearly lead across all three tasks: Gemini-3.6 achieves the best results on viewpoints (F1: 93.20) and narrative roles (F1: 64.14), while Kimi-K3 leads on social identities (F1 52.78) and ranks second on the other two tasks. The remaining models trail considerably: on viewpoints, they cluster between 71 and 80 F1, more than 13 points below the leaders, and on narrative roles, the next-best model, GLM-4.1V (F1: 46.55), scores 17.59 points below Gemini-3.6. Precision and recall reveal how this advantage is achieved. On social identities, the leaders' advantage stems from recall (R: 73.55 and 77.60), while their precision (P: 45.58 and 44.99) is comparable to that of smaller models such as Qwen3-VL-T (P: 47.99). On narrative roles, the two leaders follow different strategies: Gemini-3.6 achieves the highest precision (P: 59.82), whereas Kimi-K3 achieves the highest recall (R: 81.01) at a considerably lower precision (P: 48.11). More generally, most models show higher recall than precision on both fine-grained tasks, indicating a tendency to over-assign identities and roles rather than miss them. Full results are reported in Appendix~\ref{sec:experimental_results}, Table \ref{tab:experiment_results}.

\paragraph{$\textit{RQ}_2$: Underlying causes of model performance}
Model scale and reasoning appear to be the main factors underlying these differences, though our comparison does not control for other design choices such as training data. The two frontier-scale models, Gemini-3.6 and the sparse Mixture-of-Experts (MoE) Kimi-K3, outperform all other models on every task, with the largest margins on narrative roles (64.14 and 57.88 vs.\ at most 46.55). This cannot be explained by architecture alone: DeepSeek-VL2, also an MoE model, is one of the weakest models on social identities (13.07). Among the 7--12B models, viewpoint performance is relatively similar (71.08--79.61), whereas social identity performance varies widely (5.80--43.42). This suggests that smaller models differ in their ability to map memes onto large, fine-grained label spaces. Within this group, reasoning improves performance: the two reasoning models, Qwen3-VL-T and GLM-4.1V, are the strongest on social identities and narrative roles, and Qwen3-VL-T outperforms its instruct counterpart, on all three tasks, most notably on social identities (43.42 vs.\ 36.35). The effect of reasoning differs by task: on social identities and viewpoints it mainly increases precision (P: 47.99 vs.\ 39.66; 80.51 vs.\ 76.11), whereas on narrative roles it mainly increases recall (R: 58.57 vs.\ 55.53).

\paragraph{$\textit{RQ}_3$: Influence of frequency on performance}
Model performance correlates positively with label frequency for social identities and narrative roles, whereas template frequency shows no meaningful relationship with performance. For social identities, the correlation is statistically significant (Spearman's $\rho = 0.35$, pooled across models), with rare identities scoring considerably lower than frequent ones (e.g., women: $N = 290$, F1 = 65.5; authoritarian: $N = 1$, F1 = 3.1). Narrative roles show a similar dependence on frequency ($\rho = 0.29$), with frequent roles outperforming less common ones (e.g., villain: $N = 362$, F1 = 62.4; hero: $N = 46$, F1 = 27.2). Frequency alone, however, does not explain role performance: \textit{loser} achieves only 33.8 F1 despite its high frequency ($N = 257$), suggesting that some roles are inherently harder to interpret. In contrast, template frequency is uncorrelated with performance ($\rho = -0.01$). See Appendix~\ref{sec:experimental_results}, Figure~\ref{fig:frequency_performance} for more details.

\begin{figure}[t]
\centering
\includegraphics[width=0.8\linewidth]{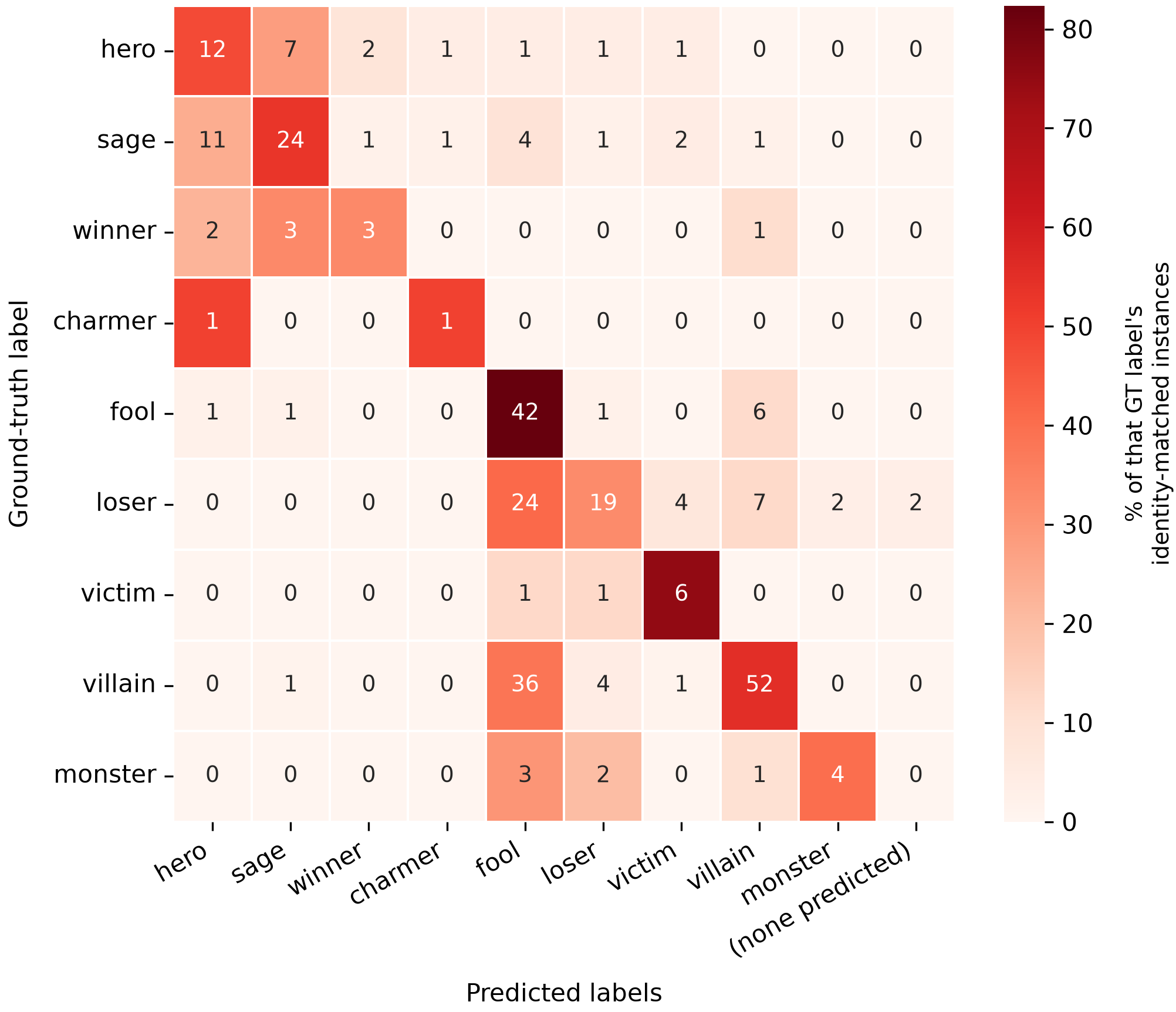}
\caption{Narrative role error analysis heatmap.}% represented as a tiled heatmap.}
\label{fig:error_analysis_gemini}
\end{figure}

\paragraph{$\textit{RQ}_4$: Impact of template knowledge on performance}
We hypothesized that knowledge of a template's conventional meaning and usage would help models infer narrative roles. To test this, we evaluated the best-performing model, Gemini-3.6, with and without template descriptions in the prompt (Table~\ref{tab:template_ablation}). Contrary to our hypothesis, adding template information reduced performance on all three metrics, with Macro-F1 dropping by 3.3 points (from 65.2 to 61.9). The largest decrease was in precision (4.9 points), while recall changed less (1.6 points). These findings indicate that, in practice, narrative role association frequently extends beyond conventional template uses, as illustrated by the examples provided in Appendix \ref{sec:ablation_study}.

\paragraph{Error analysis. } 

To compare model and human performance directly, we evaluate our best-performing model, Gemini-3.6, alongside five human evaluators on a shared subset of 214 memes without access to ground-truth labels. In this setting, models and humans must assign viewpoints and narrative roles to their own predicted identities rather than gold identities. As shown in Table~\ref{tab:gemini_vs_human}, Gemini-3.6 outperforms untrained human evaluators across all three tasks, achieving an overall average Macro F1 of 58.67\% compared to 49.38\% for humans: a gap driven by the model's superior recall in identifying social identities and narrative roles. Qualitative analysis reveals that Gemini-3.6 shares several systematic error patterns with human annotators: both tend to default to the outgroup, the majority viewpoint class, and both generally preserve the overall polarity of a portrayal (positive vs. negative) even when assigning the wrong role. However, the model’s errors are more concentrated: it frequently collapses distinct negative roles into a single category (villain), and it swaps roles between antagonists, assigning a role to the wrong identity \ref{fig:narrative_roles_error_matrix}. Human annotators, by contrast, are more likely to leave a role unassigned. Appendix~\ref{sec:error_analysis} provides a detailed breakdown of these error patterns.

% Social identity and narrative role classification yield lower scores overall, reflecting the inherent complexity of mapping inputs to a larger, more fine-grained label space. This is illustrated in Figure \ref{fig:model_comparison_heatmap}. These findings suggest that capturing abstract concepts requires high-level semantic representation that standard VLMs struggle to model without explicit reasoning pipelines or massive parameter scale. 

\begin{table}[t]
\centering
\footnotesize
\renewcommand{\arraystretch}{0.85}
\caption{Macro-averaged narrative role performance of Gemini-3.6 with and without template knowledge.}
\label{tab:template_ablation}
\begin{tabular*}{\columnwidth}{@{\extracolsep{\fill}} lccc @{}}
\toprule
\textbf{Prompt Condition} & \textbf{Precision} & \textbf{Recall} & \textbf{F1} \\
\midrule
Standard  & \bfseries 60.50 & \bfseries 77.10 & \bfseries 65.20 \\
Templated & 55.60 & 75.50 & 61.90 \\
\bottomrule
\end{tabular*}
\end{table}

\section{Conclusions}

We present IRON-Meme, a benchmark for evaluating VLM classification of social identities, viewpoints, and narrative roles in memes. Constructed using a novel framework, IRON-Meme encompasses these categories in three distinct classification tasks. An evaluation of nine VLMs demonstrates that scalable narrative analysis is attainable but remains challenging. The most effective models reliably distinguish ingroups from outgroups, but accurately identifying portrayed individuals and their roles remains complex, especially for less frequent labels. Reasoning-based models outperform instruction-based counterparts. Contrary to initial expectations, explicit template knowledge did not improve role inference, and instead reduced performance. This finding indicates that effective meme understanding goes far beyond templates' conventional use. Future research should expand IRON-Meme to include a broader range of sociopolitical contexts and languages, as well as larger-scale datasets with proportional representation of different ideologies. Additionally, further investigation is warranted into methods for integrating template knowledge without compromising meme-specific meaning, thereby advancing tools for tracking the construction of meme narratives at scale.

\section*{Limitations}
\label{sec:limitations}

Despite efforts to curate a balanced corpus, \textit{\textbf{the dataset exhibits specific compositional limitations across some dimensions}}. Specifically, we observe two constraints: \textbf{1. The data skewed toward conservative ideologies}. This imbalance results from the practical challenge of identifying dedicated, highly active subreddits focused on progressive or Democratic perspectives. Additionally, the dataset exhibits a significant bias due to the overrepresentation of misogynistic memes, primarily sourced from MAMI, the largest subset. This bias directly influences the analysis results and may affect model performance estimations. \textbf{2. The benchmark relies mostly on English-language memes from Reddit}, the study is centered on the US political and cultural ecosystem and fails to capture either non-Anglophone meme environments or the potentially more extreme narratives prevalent on alternative platforms such as 4chan. As a result of these sampling constraints, the prominence of specific identity–role combinations within the dataset reflects our collected corpus rather than online discourse as a whole. While our findings highlight meaningful patterns of identity construction within this specific context, future research should expand dataset boundaries to incorporate broader platform perspectives, non-English environments, and a more balanced representation of online political debates.

Another significant limitation of this study arises from the \textit{\textbf{inherently subjective nature of political meme interpretation}}, which is influenced by the context-dependent nature of visual tropes and the annotators' individual background beliefs. As the first empirical investigation of political meme interpretation at this scale, we encountered these complexities firsthand. Examples of these challenges are annotators assuming that voters perceive themselves as morally superior to non-voters, or reading a brain graphic directly as a symbol of intelligence. To address these challenges, \textbf{we established comprehensive annotation guidelines to facilitate the understanding of complex contextual cues}. Although some annotator disagreement persists, underscoring the intrinsic difficulty of the task and indicating important directions for future research, our benchmark offers a valuable resource for evaluating computational narrative analysis of political memes.

A lower human baseline may raise concerns regarding the quality of the benchmark, as \textit{\textbf{Gemini-3.6-flash consistently outperformed human evaluators}}. This performance gap can be attributed to two primary factors: \textbf{1. Cognitive limitations.} Human evaluators are constrained by individual cognitive biases and varying interpretations. \textbf{2. Information scale.} Vision-Language Models (VLMs) utilize extensive background knowledge, which enables identification of intricate intertextual references embedded in political memes. While additional empirical investigation is required to substantiate this performance gap, these findings are consistent with the \textit{Data Perspectivism} paradigm. Instead of treating human variation as noise, data perspectivism advocates for benchmarks that explicitly incorporate the diverse perspectives and subjective interpretations of human annotators within the knowledge representation process \citep{basilePerspectivistTurnGround2023}.

Contrary to our initial hypothesis, incorporating \textit{\textbf{template-level background knowledge did not improve narrative role classification performance}}. We attribute this outcome to: \textbf{1. Meme fluidity and pragmatic use:} While templates establish baseline conventional expectations, political memes are inherently dynamic: meanings are continuously repurposed, inverted, or deployed ironically. Consequently, narrative roles depend primarily on instance-specific communicative intent rather than static template definitions. \textbf{2. Information scope:} The background information provided by platforms such as \textit{Know Your Meme} focuses largely on origins and intertextual references rather than conventional usage patterns. Future work should explore whether higher-quality or more pragmatically focused template annotations can enhance role classification.

\section*{Ethical considerations}

To support social media researchers in analyzing online discourse, this study provides a benchmark for evaluating VLM capabilities in decoding complex digital narratives. The authors explicitly state that this benchmark and its underlying framework are intended strictly for academic and research purposes; we do not authorize their deployment for automated content moderation, censorship, or commercial surveillance.

All meme content included in the benchmark has been carefully anonymized, with identifiable personal information removed. Our analytical focus is restricted to macro-level narrative patterns and identity construction within political memes. We do not aim to evaluate, target, or depict any individual social media user, nor do we endorse or promote any of the ideological stances represented within the dataset.

The research adhered to the Code of Ethics for Research Involving Human Participants established by the relevant university faculty. Compliance with ethical requirements was confirmed through a self-assessment process.

\section*{Acknowledgments}

We gratefully acknowledge the annotators Letizia Abbate, Milan van Wordragen, and Jana Morgagni, whose contributions were essential to this project. We also thank the Societal Analytics Lab for awarding us the annotation grant that funded this aspect of the research (grant no. 258280). We appreciate the feedback provided by all PhD students from the Commonsense AI group on earlier versions of the paper, especially Mohammadhossein Khojasteh and Shaurya Gaur for their valuable contributions and insights. This research is supported by the NWO AiNed project “Human-Centric AI Agents with Common Sense” (grant no. NGF.1607.22.044).

\paragraph{Use of AI Assistants} We employed Claude Code to support with coding and utilized Grammarly and Gemini to refine the manuscript. All methodological and analytical decisions were made by the authors.
% ...


\bibliography{references}

\clearpage

\appendix

\twocolumn[{%
  \section{Experimental details}
  \label{sec:appendix_a}
  \subsection{Annotation interface}
  \label{sec:annotation_interface}

  \centering
  \captionsetup{type=figure}
    \includegraphics[width=\textwidth,height=0.75\textheight,keepaspectratio]{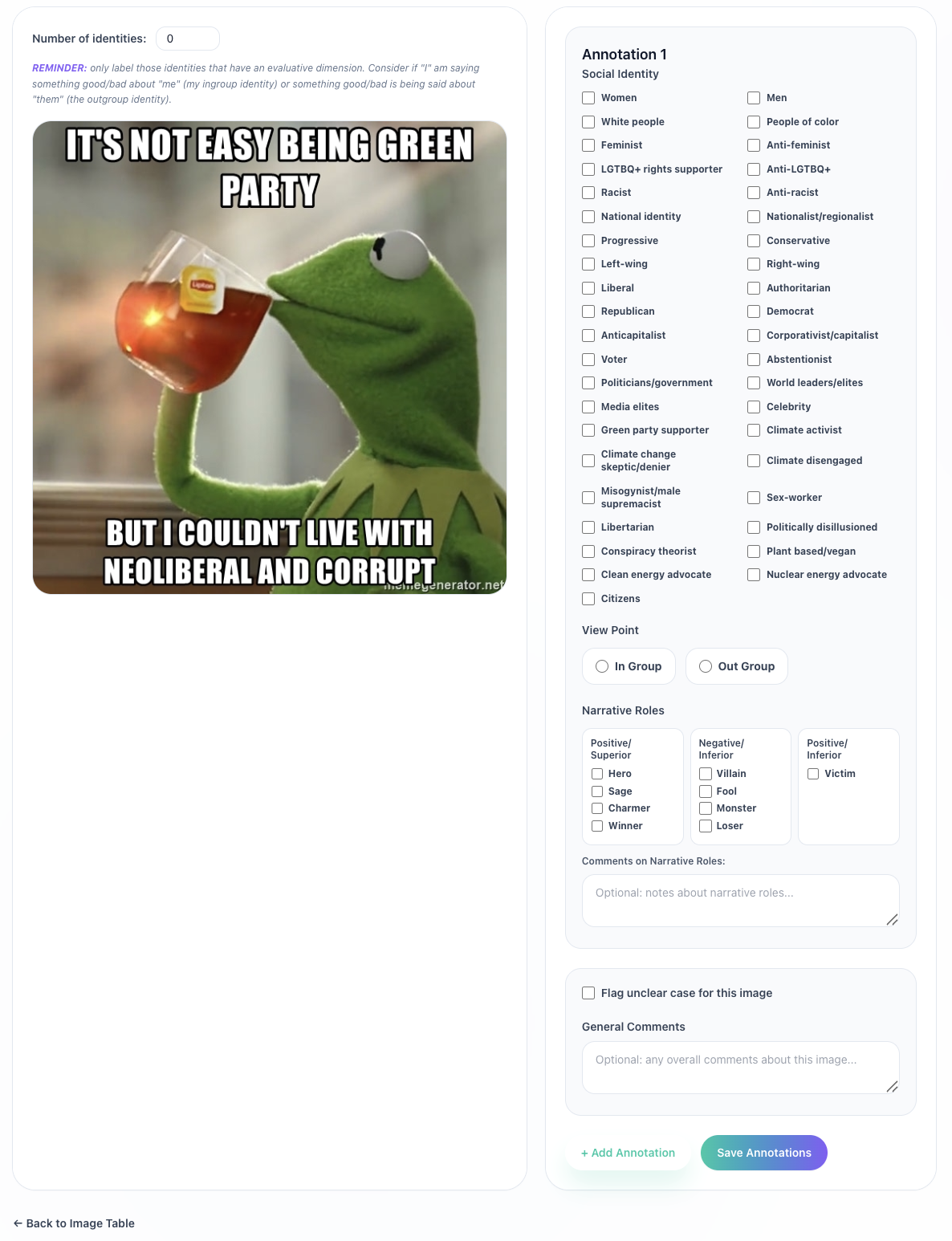}
  \caption{\textbf{Screenshot of the annotation interface used by IRON-Meme annotators.} For each annotation record, annotators select one identity, one viewpoint, and one or more narrative roles. To add another social identity to a meme, the \textit{+ Add annotation} button is used. The interface also provides options to flag unclear cases for later review and to add comments regarding the meme or the annotations.}
  \label{fig:annotation_interface}
  \vspace{1em}
}]

\clearpage
\FloatBarrier

\subsection{Social identity labels}
\label{sec:social_identity_labels}

Table \ref{tab:identity_labels} details the social identity labels evaluated, spanning demographics (e.g., gender, ethnicity) as well as general and domain-specific ideological positions.

\begin{table}[ht]
\centering
\scriptsize
\setlength{\tabcolsep}{4pt}
\renewcommand{\arraystretch}{1.2}
\caption{Set of social identity labels used in the dataset.} 
\label{tab:identity_labels}

\resizebox{\columnwidth}{!}{%
\begin{tabular}{lll}
\toprule
\multicolumn{3}{c}{\scriptsize\textbf{Identity Label}} \\
\midrule
Women & Men & White people \\
People of color & Citizens & National identity \\
Feminist & Anti-feminist & Sex worker \\
LGBTQ+ rights supporter & Anti-LGBTQ+ & Misogynist/male supremacist \\
Racist & Anti-racist & Nationalist/regionalist \\
Progressive & Conservative & Liberal \\
Left-wing & Right-wing & Libertarian \\
Authoritarian & Democrat & Republican \\
Anticapitalist & Corporatist/capitalist & Green Party supporter \\
Voter & Abstentionist & Politically disillusioned \\
Politicians/government & World leaders/elites & Media elites \\
Celebrity & Conspiracy theorist & Climate activist \\
Climate change skeptic/denier & Climate disengaged & Clean energy advocate \\
Nuclear energy advocate & Plant-based/vegan & \\
\bottomrule
\end{tabular}%
}
\end{table}

\subsection{Computational resources}

All open-weight models (i.e., DeepSeek-VL2, GLM-4.1V-9B-Thinking, Molmo-7B-D-0924, Pixtral-12B-2409, Qwen2.5-VL-7B-Instruct, Qwen3-VL-8B-Instruct, and Qwen3-VL-8B-Thinking) were served locally with vLLM on a SLURM cluster partition. Each job was allocated one NVIDIA H100 GPU (94 GB), 8 CPU cores, 48 GB of RAM, and a 12-hour wall-time limit. The sole exception was Kimi-K3 (a 2.8T-parameter MoE model with ~104B active parameters), which was evaluated through Moonshot's hosted, OpenAI-compatible API, because its MXFP4-quantized weights (approximately 1.4 TB) far exceed the memory of a single GPU. Since the API only accepts a sampling temperature of 1, Kimi-K3 was decoded non-greedily, unlike the locally served models (temperature 0).

\subsection{Parsing}
\label{sec:parsing}

Although prompts request JSON output, model generation is unconstrained. Responses are therefore parsed by stripping reasoning traces and code fences before extracting the outermost JSON span; unparseable responses are scored as empty predictions. Predicted labels are normalized across case, punctuation and separators, pluralization, and spelling variants, and retained only if they exist in the corresponding task's label inventory; all other labels are discarded. Evaluation metrics are computed over each task's fixed label inventory, so discarded labels do not introduce additional classes. Table~\ref{tab:unparseable_rates} reports parsing failures and Table~\ref{tab:discarded_rates} reports discarded labels per model.

Evaluation diagnostics identified distinct failure modes among models during parsing and label validation. Molmo (molmo-7b-d-0924) exhibited significant degradation, failing to generate parseable JSON in over half (52.7\%) of its step-1 responses and having two-thirds of its predicted role labels discarded. These issues primarily resulted from producing Chinese translations, substituting viewpoint terms (outgroup/ingroup, approximately 1,700 instances), or altering dimension names (e.g., Moral). Other models primarily failed due to prompt leakage or schema mismatches. For instance, deepseek-vl2 frequently repeated prompt example mappings as identity labels (e.g., “Extinction Rebellion”, “Neoliberal”) and substituted viewpoint terms for roles. Qwen3-8b-thinking copied prompt placeholder tokens (IDENTITY\_1, IDENTITY\_2) verbatim. Several models predicted valid domain concepts that were not included in the strict task inventory, such as qwen2.5-7b-instruct using dimension descriptors like Inferior/Positive as roles, or glm-4.1v-9b-thinking generating unmapped identities such as “Government” or “Girls”. Additionally, minor inflection mismatches, such as predicting “Woman” instead of the approved plural label “Women”, were filtered out because standard normalization rules only removed trailing “s” characters. This highlights the need to expand strict spelling aliases to accommodate valid morphological variants.

\begin{table}[htbp]
\centering
\caption{Percentage of unparseable responses per model across task steps.}
\label{tab:unparseable_rates}
\scriptsize
\setlength{\tabcolsep}{8pt}
\begin{tabular}{lccc}
\toprule
\textbf{Model} & \textbf{s1 (Ident.)} & \textbf{s2 (View.)} & \textbf{s3 (Roles)} \\
\midrule
deepseek-vl2 & 10.2\% & 1.4\% & 0.0\% \\
gemini-3.6-flash$\dagger$ & — & — & — \\
glm-4.1v-9b-think & 1.0\% & 1.5\% & 0.0\% \\
kimi-k3 & 0.0\% & 0.0\% & 0.0\% \\
molmo-7b-d & 52.7\% & 2.3\% & 1.4\% \\
pixtral-12b & 0.0\% & 0.0\% & 0.0\% \\
qwen2.5-vl-7b & 0.0\% & 0.0\% & 0.0\% \\
qwen3-vl-8b & 0.0\% & 0.0\% & 0.0\% \\
qwen3-vl-8b-think & 3.1\% & 6.0\% & 5.0\% \\
\bottomrule
\end{tabular}
\end{table}

\begin{table}[htbp]
\centering
\caption{Percentage of discarded labels outside task inventories per model.}
\label{tab:discarded_rates}
\scriptsize
\setlength{\tabcolsep}{8pt}
\begin{tabular}{lccc}
\toprule
\textbf{Model} & \textbf{Identities} & \textbf{Viewpoints} & \textbf{Roles} \\
\midrule
deepseek-vl2 & 7.2\% & 0.0\% & 5.2\% \\
gemini-3.6-flash$\dagger$ & 0.0\% & 0.0\% & — \\
glm-4.1v-9b-think & 1.3\% & 0.0\% & 0.0\% \\
kimi-k3 & 0.0\% & 0.0\% & 0.0\% \\
molmo-7b-d & 47.7\% & 0.2\% & 0.0\% \\
pixtral-12b & 3.4\% & 0.0\% & 0.0\% \\
qwen2.5-vl-7b & 3.4\% & 0.0\% & 0.6\% \\
qwen3-vl-8b & 0.4\% & 0.0\% & $<0.1\%$ \\
qwen3-vl-8b-think & 0.2\% & — & — \\
\bottomrule
\end{tabular}
\end{table}

\subsection{Prompts}

\captionsetup[lstlisting]{labelformat=simple}

\label{sec:prompts}
\begin{lstlisting}[
  basicstyle=\ttfamily\small,
  breaklines=true,
  frame=none,
  numbers=none,
  caption={Prompt for Social Identity Extraction},
  label={lst:prompt_social_identity}
]
SYSTEM: You are an expert in semiotics and cultural analysis. Analyze memes to identify evaluatively represented social identities.
Return valid JSON only. Do not include markdown, explanations, or text outside the JSON object.

USER: Analyze the attached meme. Return:
{"social_identities":["IDENTITY_1","IDENTITY_2"]}
Select identity labels ONLY from the approved list below.

RULES
- Label an identity only if the meme expresses or strongly implies an evaluative stance toward it.
- Do not make up custom labels, sub-categories, or descriptive phrases (e.g., do NOT output "angry leftists", "neoliberals", or "communists"). Use exact labels from the approved list.

MAPPING & ABSTRACTION RULES
If a specific sub-ideology, political movement, or granular social group is depicted or implied, map it to its broader, general umbrella label from the approved list.
Examples:
- Communist / Socialist / Marxist / Anarchist -> "Left wing" or "Anticapitalist"
- Neoliberal / Classical Liberal / Capitalist -> "Liberal" or "Corporativist/capitalist"
- Fascist / Alt-Right / Far-Right -> "Right-wing" or "Authoritarian"
- Neoconservative / Alt-Lite -> "Conservative"
- Climate Striker / Extinction Rebellion / Eco-activist -> "Climate activist"
- Queer / Transgender / Gay / Lesbian -> "LGTBQ+ rights supporter" or "Anti-LGTBQ+" (depending on the stance/group evaluated)

APPROVED IDENTITY LABELS:
[
  "Women",
  "Men",
  "White people",
  "Feminist",
  "Anti feminist",
  "LGTBQ+ rights supporter",
  "Anti LGTBQ+",
  "Misogynist/male supremacist",
  "Sex worker",
  "Racist",
  "Anti racist",
  "people of color",
  "National identity",
  "Nationalist/regionalist",
  "Progressive",
  "Conservative",
  "Left wing",
  "Right wing",
  "Liberal",
  "Authoritarian",
  "Republican",
  "Democrat",
  "Anticapitalist",
  "Corporativist/capitalist",
  "Voter",
  "Abstentionist",
  "Politicians/government",
  "World leaders/elites",
  "Media elites",
  "Celebrity",
  "Green party supporter",
  "Libertarian",
  "Politically disillusioned",
  "Conspiracy theorist",
  "Climate activist",
  "Climate change skeptic/denier",
  "Climate disengaged",
  "Plant based/vegan",
  "Clean energy advocate",
  "Nuclear energy advocate",
  "Citizens"
]

OUTPUT REQUIREMENTS
Output valid JSON only. Key must be "social_identities". Value must be an array of approved strings strictly matching the list above.
\end{lstlisting}

\begin{lstlisting}[
  basicstyle=\ttfamily\small,
  breaklines=true,
  frame=none,
  numbers=none,
  caption={Prompt for Viewpoint Classification},
  label={lst:prompt_viewpoint}
]
SYSTEM: You are an expert in semiotics and cultural analysis specializing in meme interpretation. Your task is to identify the viewpoint expressed by a meme relative to the social identities identified in the meme.
Return valid JSON only. Do not include markdown, explanations, commentary, or any text outside the JSON object.

USER: Analyze the attached meme and determine the viewpoint expressed by the meme toward ALL of the following identified social identities: {social_identities}.

STRICT INSTRUCTION:
You MUST evaluate and return a viewpoint entry for EVERY single identity listed in {social_identities}. Do not omit any identity.

Return JSON array where each entry specifies an identity and its viewpoint:
[
  {"identity": "IDENTITY_NAME", "viewpoint": "ingroup"},
  {"identity": "IDENTITY_NAME", "viewpoint": "outgroup"}
]

VIEWPOINT DEFINITION
The ingroup is the social identity whose perspective, beliefs, values, actions, or position are expressed, supported, normalized, defended, or endorsed by the meme.
The outgroup is the social identity that is criticized, mocked, blamed, rejected, attacked, negatively portrayed, or positioned as opposed to the meme's perspective.

OUTPUT REQUIREMENTS
Output valid JSON array only.
Length of output array MUST equal the number of input identities provided.
Every entity MUST have a 'viewpoint' field.
Each object must have exactly two keys: "identity" and "viewpoint".
Each "identity" key must have only one "viewpoint" assigned.
"viewpoint" must be either "ingroup" or "outgroup".
\end{lstlisting}

\begin{lstlisting}[
  basicstyle=\ttfamily\small,
  breaklines=true,
  frame=none,
  numbers=none,
  caption={Prompt for Narrative Role Classification},
  label={lst:prompt_narrative_role}
]
SYSTEM: You are an expert in semiotics and cultural analysis specializing in meme interpretation. Your task is to identify the narrative roles assigned to the social identities present in a meme.
Return valid JSON only. Do not include markdown, explanations, commentary, or any text outside the JSON object.

USER: Analyze the attached meme and assign narrative roles to the previously identified social identities and viewpoints:
{viewpoints_context}

Narrative roles describe how a social identity is portrayed within the meme's message. Assign roles relative to the identity's representation, evaluation, and position in the narrative.

Return a JSON array formatted as follows:
[
  {
    "identity": "IDENTITY_NAME",
    "roles": ["ROLE_1", "ROLE_2"]
  }
]

ROLE FRAMEWORK
Follow two steps:
Step 1: Identify the evaluative dimension (Moral, Intellectual, Physical, Power)
Step 2: Determine whether the portrayal is Superior/Positive or Inferior/Negative.

NARRATIVE ROLES
Moral dimension:
- Hero (Superior/Positive): Morally good, ethical, justified, virtuous, protective.
- Victim (Inferior/Positive): Harmed, oppressed, unfairly treated, vulnerable, deserving sympathy. Note that victims usually correspond to ingroup viewpoints.
- Villain (Inferior/Negative): Morally bad, harmful, unethical, unjustified, threatening.

Intellectual dimension:
- Sage (Superior/Positive): Knowledgeable, rational, intelligent, insightful, informed.
- Fool (Inferior/Negative): Ignorant, irrational, uninformed, incompetent, mistaken.

Physical dimension:
- Charmer (Superior/Positive): Attractive, appealing, admirable, charismatic.
- Monster (Inferior/Negative): Repulsive, threatening, undesirable, dehumanized.

Power dimension:
- Winner (Superior/Positive): Successful, dominant, influential, empowered, victorious.
- Loser (Inferior/Negative): Defeated, powerless, unsuccessful, marginalized.

ANNOTATION RULES
- Assign roles to social identities, not individuals.
- An identity must have exactly 1 viewpoint (if provided) and can receive 1 or many narrative roles (e.g., ["Hero", "Winner"]).
- Assign roles only when supported by the meme's text, imagery, or framing.
- If an identity has no clear narrative role, return an empty array for roles.
\end{lstlisting}

\begin{figure*}[ht]
\centering
\makebox[\textwidth][c]{\includegraphics[width=0.9\textwidth]{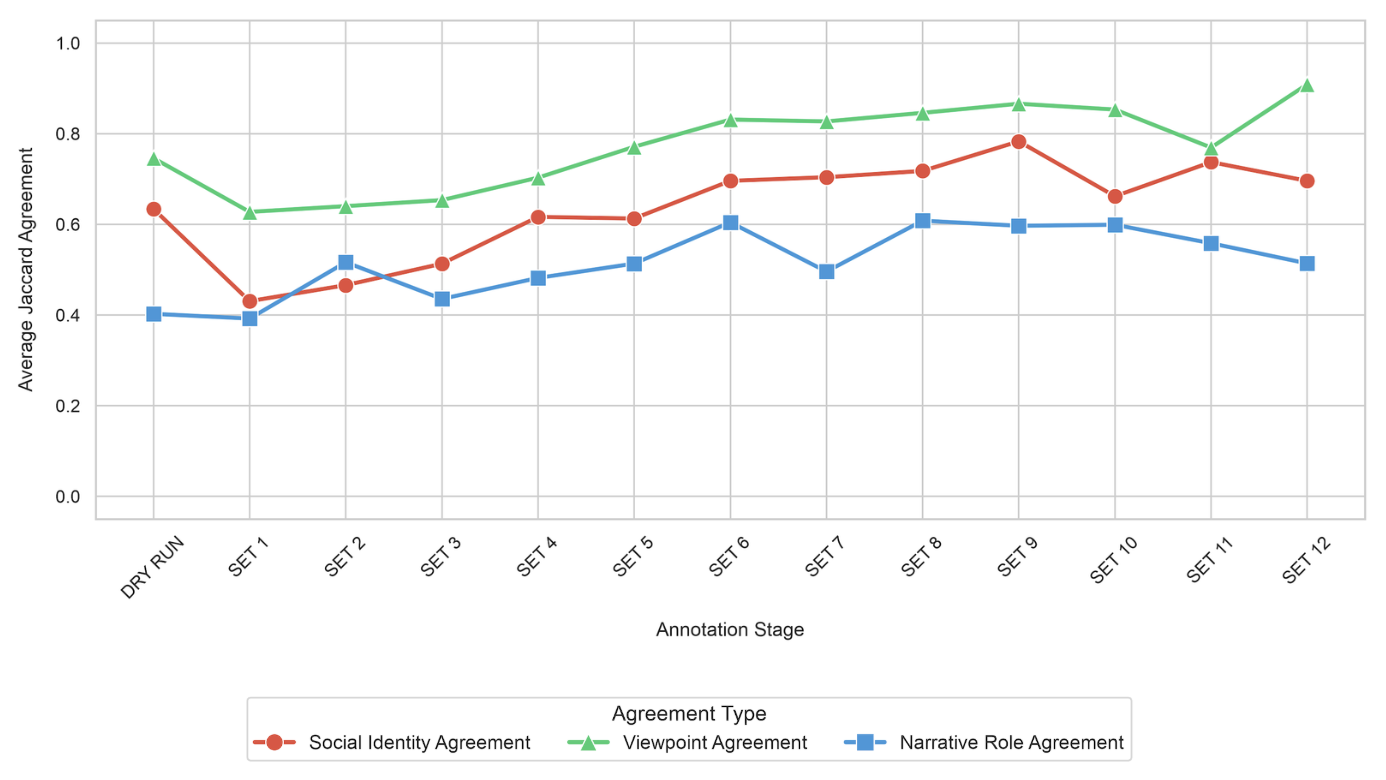}}
\caption{Pairwise Jaccard agreement across annotation stages}
\label{fig:jaccard_agreement_comparison}
\end{figure*}

\subsection{Agreement metrics}
\label{sec:agreement_metrics}
Complete inter-rater agreement scores across all categories are summarized in Table~\ref{tab:inter_rater_agreement}. Ground-truth labels were established by retaining annotations agreed upon by at least two raters across $1,110$ final memes ($78.87\%$ retention rate out of $1,410$ total annotated; $30.85\%$ perfect agreement, $48.01\%$ majority consensus). 
Social identity emerged as the most reliably annotated category, exhibiting the highest Fleiss' and Cohen's $\kappa$. The close correspondence between these two $\kappa$ measures demonstrates that agreement is consistent across all three raters, rather than being influenced by a specific rater pair. Viewpoint annotation yielded a higher Jaccard score and rate of perfect agreement, but lower $\kappa$ values, indicating that raters generally identified overlapping viewpoints while showing less consistent agreement beyond exact matches. Agreement was weakest for narrative-role classification, suggesting that this category is more subjective or challenging to operationalize, likely due to the need for greater contextual interpretation. Nevertheless, $\kappa$ values of approximately 0.51 reflect moderate inter-rater agreement. In summary, agreement decreases as the annotation task becomes more relational and compositional, with the lowest agreement observed for the combined relational identity, viewpoint, and role triples.

Figure \ref{fig:jaccard_agreement_comparison} shows changes in annotator agreement across successive annotation stages. Agreement generally increases through SET 6, coinciding with the period during which the annotation guidelines and codebook were iteratively refined. After SET 6, agreement largely stabilizes, with some fluctuations across subsequent stages. This pattern suggests that the refinement of the guidelines contributed to improved annotation consistency, while the subsequent stabilization indicates increasing convergence among annotators.

\begin{table}[!b]
\centering
\footnotesize
\setlength{\tabcolsep}{3pt} 
\renewcommand{\arraystretch}{1.15}
\begin{threeparttable}
\caption{Inter-rater agreement metrics across annotation categories.}
\label{tab:inter_rater_agreement}
\begin{tabularx}{\linewidth}{@{} X *{4}{c} @{}}
\toprule
\textbf{Category} & \textbf{Fleiss' $\kappa$\tnote{a}} & \textbf{Avg. $\kappa$\tnote{b}} & \textbf{Jaccard} & \textbf{Perfect} \\
\midrule
Social identity & 0.694 & 0.693 & 0.660 & 45.82\% \\
Viewpoint       & 0.594 & 0.593 & 0.794 & 60.07\% \\
Narrative roles & 0.509 & 0.508 & 0.532 & 26.31\% \\
\midrule
Overall (Triples)\tnote{c} & 0.426 & 0.425 & 0.403 & 30.85\% \\
\bottomrule
\end{tabularx}
\begin{tablenotes}
  \small
  \item[a] Evaluated across 3 raters. Fleiss' $\kappa > 0.60$ indicates substantial agreement; $0.41$--$0.60$ indicates moderate agreement.
  \item[b] Average pairwise Cohen's $\kappa$.
  \item[c] Combined relational identity--viewpoint--role triples.
\end{tablenotes}
\end{threeparttable}
\end{table}

\subsection{Template-based filtering}
\label{sec:template_based_filtering}

\begin{figure*}[ht]
\centering
\footnotesize
\setlength{\tabcolsep}{3pt} 
\renewcommand{\arraystretch}{1.15}

\includegraphics[width=0.9\linewidth]{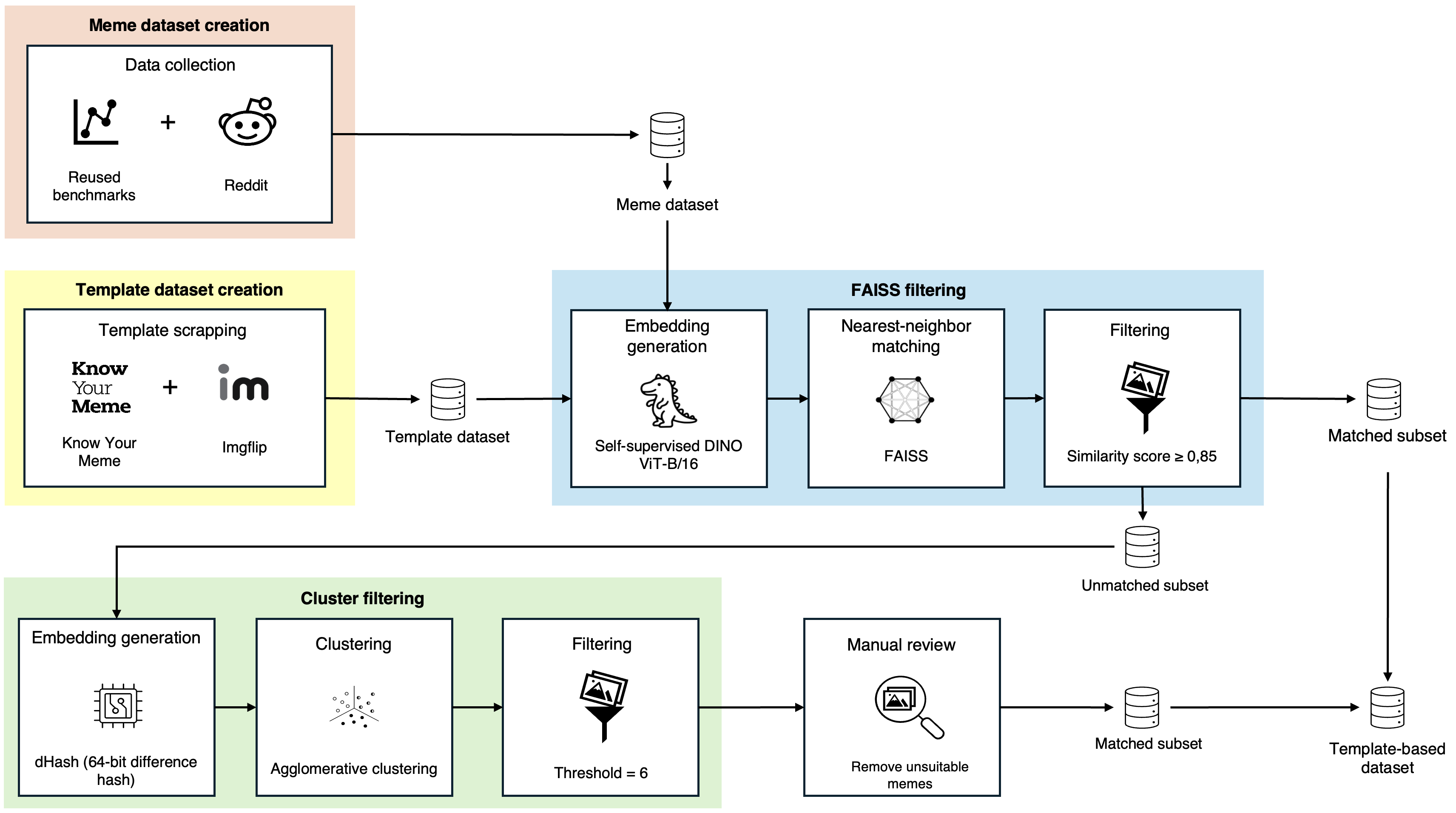}
\caption{Overview of the template-based meme curation pipeline.} 
\label{fig:template_based_filtering}
\end{figure*}

Only memes corresponding to established templates are retained to reduce ambiguity in communicative functions due to their structured and formalizable nature. The filtering pipeline is detailed in Figure \ref{fig:template_based_filtering}. \paragraph{Data creation} A template dataset of 5,177 templates was constructed by scraping Know Your Meme (4,703) and Imgflip (474). \paragraph{FAISS filtering} 384-dimensional visual embeddings were generated for both memes and templates using a self-supervised DINO ViT-B/16 model, with all embeddings L2-normalized to measure similarity as cosine similarity. Exact nearest-neighbor matching was performed via FAISS \citep{douzeFaissLibrary2025}, filtering matched memes with a top similarity score $\ge 0.85$. \paragraph{Cluster filtering} Unmatched memes were evaluated for near-duplicates using a 64-bit difference hash (dHash) and grouped via agglomerative clustering with a distance threshold of 6. \paragraph{Manual review} Manual inspection removed unsuccessful matches, non-memes, non-template memes, and unsuitable content. This process yields a dataset of 1,410 memes.

\subsection{Metrics}
\label{sec:metrics}

Let C denote the total number of classes in the evaluated task, and let c represent an individual class (social identity, viewpoint, or narrative role), where $c \in \{1, ..., C\}$. For each class c, TP\_c, FP\_c, and FN\_c indicate the numbers of true-positive, false-positive, and false-negative predictions, respectively. Micro-averaged metrics aggregate the contributions of all classes prior to metric calculation, thereby giving greater influence to classes with more instances. In contrast, macro-averaged metrics are computed independently for each class and then averaged across all classes, assigning equal weight to each class regardless of its frequency. The corresponding precision, recall, and F1-scores are defined as follows:

{\small
\begin{equation}
F1_{\mathrm{micro}} =
\frac{2\sum_{c=1}^{C} TP_c}
{2\sum_{c=1}^{C} TP_c + \sum_{c=1}^{C} FP_c + \sum_{c=1}^{C} FN_c},
\end{equation}

\begin{equation}
F1_{\mathrm{macro}} =
\frac{1}{C}\sum_{c=1}^{C}
\frac{2TP_c}{2TP_c + FP_c + FN_c}.
\end{equation}
}

\subsection{Benchmark analysis}
\label{sec:benchmark_analysis}

\begin{figure}[b]
  \centering
  \includegraphics[width=\linewidth]{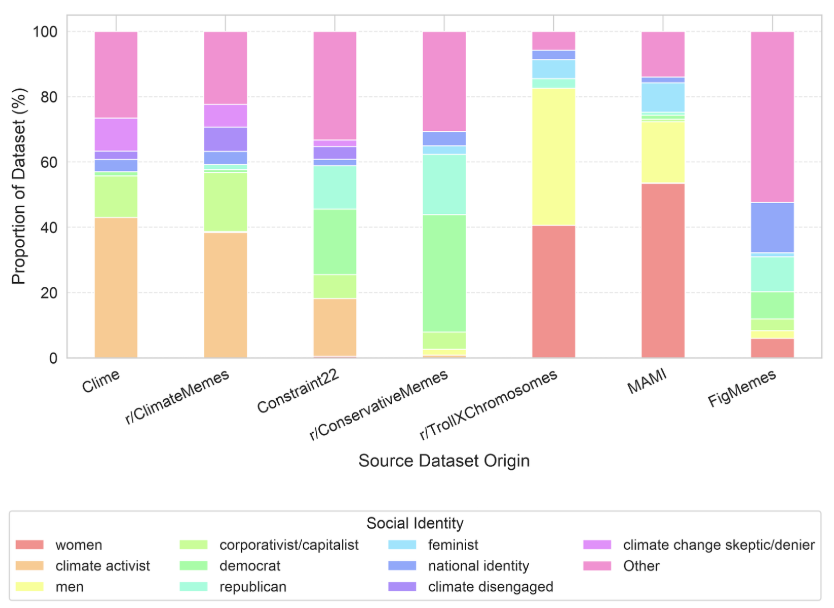}
  \caption{Distribution of Social Identities Across Source Sub-Datasets.} 
  \label{fig:identity_by_source}
\end{figure}

Table \ref{fig:identity_by_source} shows the relative proportion (\%) of identified social identity labels across the different source datasets on the x-axis. Each stacked bar illustrates the domain-specific representation of identities. As expected, \textit{climate activist} and \textit{climate change skeptic/denier} labels are more prominent in climate-focused sub-datasets (Clime, r/ClimateMemes), \textit{women} in gender-focused sub-datasets (MAMI, r/TrollXChromosomes), and \textit{corporativist} and \textit{republican} in politically-oriented datasets (r/ConservativeMemes and Constraint22). Less expectedly, we observe a more balanced representation of \textit{women} and \textit{men} identities in r/TrollXChromosomes, focused on feminist memes, compared to the MAMI dataset, containing misogynistic memes, which contains more representations of women and fewer of men, with men serving as the primary ingroup. Similarly, MAMI contains a higher proportion of the \textit{feminist} identity. Finally, FigMemes, a dataset focused on politically-incorrect content, contains a higher proportion of \textit{national identities} than any of the other datasets.

\begin{figure}[t]
  \centering
  \includegraphics[width=\linewidth]{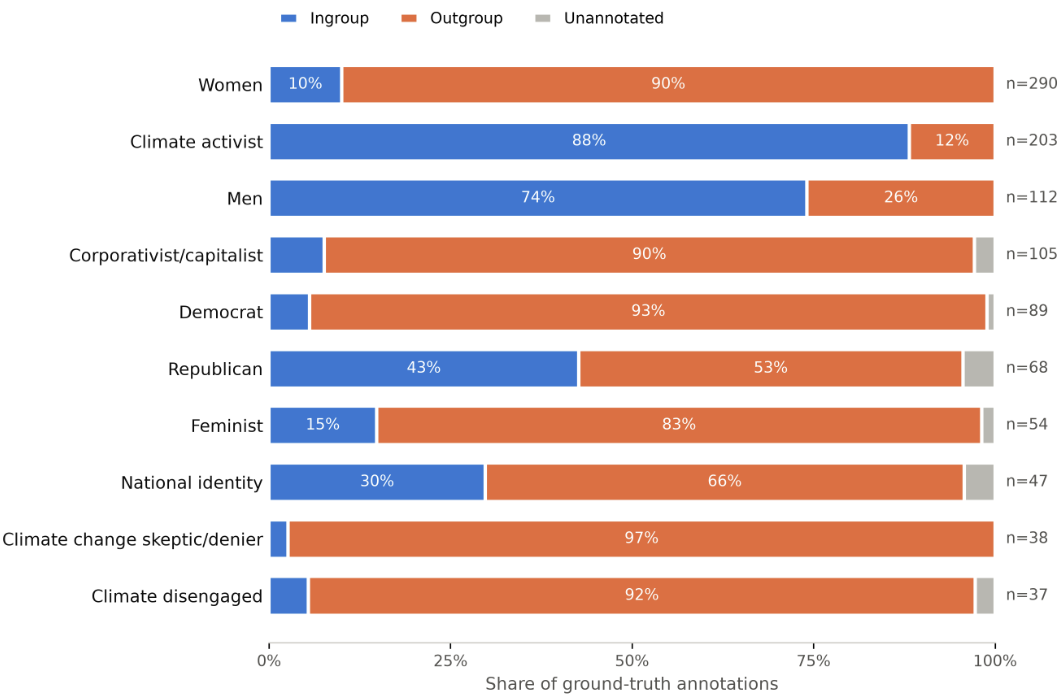}
    \caption{Viewpoints for the 10 most common social identities.}
    \label{fig:viewpoint_by_identities}
\end{figure}

Figure \ref{fig:viewpoint_by_identities} presents the distribution of viewpoints (ingroup vs. outgroup) per social identity in the IRON-Meme dataset, clarifying which identities are portrayed primarily from a sympathetic ingroup perspective versus a targeted outgroup one. The dataset exhibits a stark asymmetry across identity categories: over 80\% of representations for \textit{women}, \textit{corporativist/capitalist}, \textit{democrat}, \textit{feminist}, \textit{climate change skeptic/denier}, and \textit{climate disengaged} reflect an outgroup viewpoint, indicating that these groups serve predominantly as targets of satire, hostility, or critique. Conversely, \textit{climate activists} and \textit{men} are framed overwhelmingly from an ingroup perspective, serving as the core self-representing figures and ideological voices within the corpus. Mid-spectrum identities like \textit{republican} and \textit{national identities} show a more balanced distribution between ingroup and outgroup portrayals, pointing to active cross-ideological debate surrounding these categories. Taken together, these patterns strongly suggest that the dataset's posters are primarily male, climate-active, and anti-Democrat. Interestingly, while \textit{climate activists} and \textit{climate change skeptics/deniers} occupy opposing sides of the environmental debate, the dataset treats activists as the primary ingroup while framing skeptics almost exclusively as an outgroup; this indicates that climate-skeptical content in this dataset consists mostly of outgroup caricatures posted by climate activists rather than organic self-representation by skeptics.

Table \ref{tab:viewpoint_categories} shows the distribution of ingroup versus outgroup representations in the final IRON-Meme dataset. Outgroup representations significantly outnumber ingroup ones across the corpus. This predominance of outgroups directly aligns with our earlier finding that negative narrative roles are far more prevalent than positive ones, further confirming that meme discourse in this dataset primarily functions as a medium for targeting, critiquing, or satirizing opposing groups. 

\begin{figure}[b]
  \centering
  \includegraphics[width=\columnwidth]{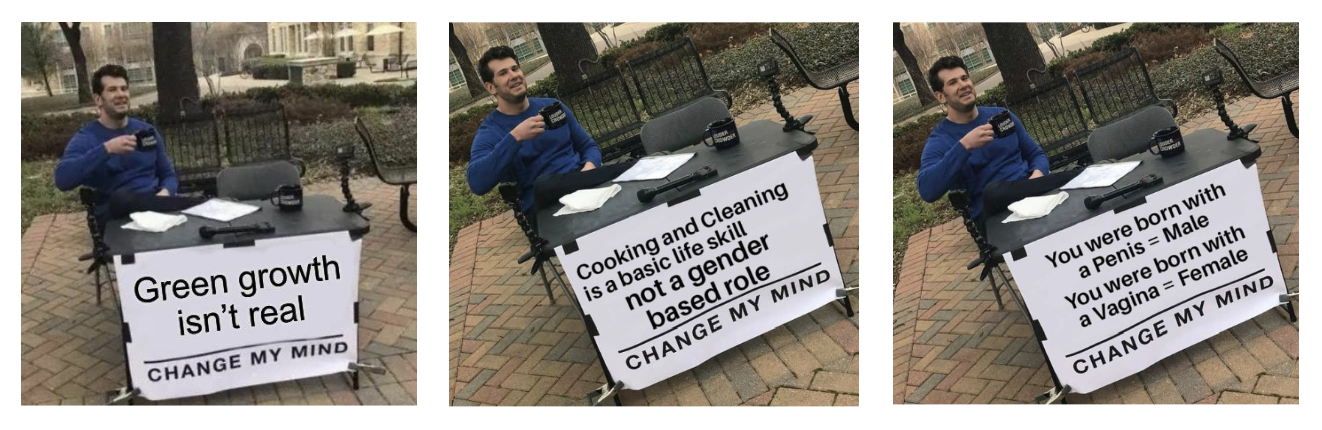}
  \caption{Examples of memes using the \textit{Steven Crowder's “Change My Mind” Campus Sign} template.}
  \label{fig:ablation_example}
\end{figure}

\begin{figure*}[ht]
\centering
\makebox[\textwidth][c]{\includegraphics[width=\textwidth]{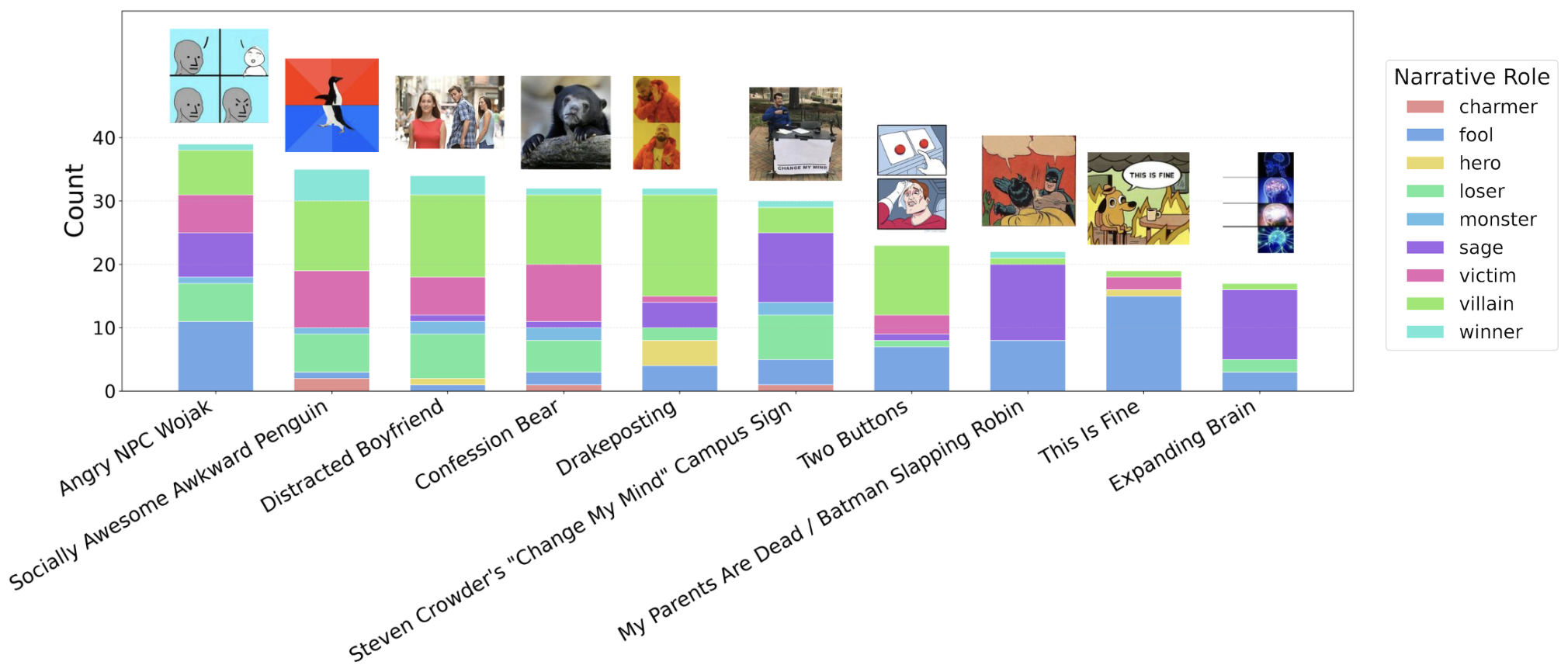}}
\caption{\textbf{Narrative-role composition for the ten most frequently used meme templates.} Stacked bars represent the number of memes assigned to each narrative role for each template. Template names correspond to their listings in \textit{Know Your Meme.}}
\label{fig:template_roles}
\end{figure*}

A key rationale for hypothesizing that template information influences narrative role classification is the observed predominance of specific role categories within certain templates. As shown in Figure \ref{fig:template_roles}, templates such as \textit{Distracted Boyfriend}, \textit{Socially Awesome Awkward Penguin}, \textit{Drakeposting}, and \textit{Confession Bear} exhibit a strong prevalence of villain and victim roles. The \textit{This Is Fine} template is almost exclusively associated with the fool role, demonstrating how a template's communicative context shapes role assignment. Formats including Steven Crowder's \textit{"Change My Mind" Campus Sign}, \textit{My Parents Are Dead / Batman Slapping Robin}, and \textit{Expanding Brain} display a substantial proportion of the sage role, which can be attributed to their conventional use in political discourse, often to present a controversial or assertive viewpoint. Furthermore, Figure \ref{fig:ablation_example} presents three examples in which annotators assigned the relevant identity the role “the sage”, suggesting that template conventions may provide cues relevant to narrative-role classification.

\begin{table}[t]
\centering
\small
\caption{Distribution of viewpoint categories.}
\label{tab:viewpoint_categories}
\begin{tabular}{lrr}
\toprule
\textbf{Category} & \textbf{Count} & \textbf{\%} \\ 
\midrule
Out-group only & 650 & 58.56 \\
In-group only & 251 & 22.61 \\
Both in- and out-group & 206 & 18.56 \\
No valid viewpoint & 3 & 0.27 \\ 
\midrule
\textbf{Total} & \textbf{1,110} & \textbf{100.00} \\ 
\bottomrule
\end{tabular}
\end{table}

\subsection{Experimental results}
\label{sec:experimental_results}

\paragraph{Model selection} The evaluation includes both reasoning-oriented models (GLM-4.1V-9B-Thinking, Qwen3-VL-8B-Thinking) and instruction-oriented models (Qwen2.5-VL-7B-Instruct, Qwen3-VL-8B-Instruct) to assess whether explicit reasoning capabilities confer advantages on tasks involving socially and narratively complex visual content. Models with varying parameter scales are included, ranging from compact models in the 7–12B range (Molmo-7B, Qwen2.5-VL-7B, Qwen3-VL-8B, GLM-4.1V-9B, Pixtral-12B) to the substantially larger Kimi-K3, which serves as a high-capacity reference point for evaluating the impact of model scale on benchmark performance. Multiple model families (DeepSeek, GLM, Kimi, Molmo, Pixtral, and Qwen) were represented to prevent results from being attributed to a single architecture or training pipeline. Additionally, the selection encompasses geographic diversity, with models developed by research organizations in East Asia (Kimi-K3, GLM-4.1V, Qwen2.5-VL, Qwen3-VL), Europe (Pixtral), and North America (Molmo), a particularly relevant dimension for a socially grounded benchmark, as differences in training data, alignment procedures, and development contexts may influence model behavior.

\begin{table*}[t]
\centering
\footnotesize
\renewcommand{\arraystretch}{0.85}
\caption{Overview of the VLMs selected for evaluation, covering diverse model families,
architectures, reasoning orientations, parameter scales, and geographic regions.}
\label{tab:vlm_details}
\begin{tabular}{llccc}
\toprule
\textbf{Model} & \textbf{Developer / Region} & \textbf{Family} & \textbf{Orientation / Capability} & \textbf{Parameter Scale} \\
\midrule
DeepSeek-VL2 & DeepSeek-AI (East Asia) & DeepSeek & Instruction & 27.5B (MoE) \\
Molmo-7B & Allen Institute for AI (North America) & Molmo & Instruction & 7B \\
Qwen2.5-VL-7B-Instruct & Alibaba Cloud (East Asia) & Qwen & Instruction & 7B \\
Qwen3-VL-8B-Instruct & Alibaba Cloud (East Asia) & Qwen & Instruction & 8B \\
Qwen3-VL-8B-Thinking & Alibaba Cloud (East Asia) & Qwen & Explicit Reasoning & 8B \\
GLM-4.1V-9B-Thinking & Zhipu AI / Tsinghua (East Asia) & GLM & Explicit Reasoning & 9B \\
Pixtral-12B & Mistral AI (Europe) & Pixtral & Instruction / Standard & 12B \\
Gemini-3.6-Flash & Google (North America) & Gemini & High-Efficiency / Reasoning & Proprietary \\
Kimi-K3 & Moonshot AI (East Asia) & Kimi & High-Capacity Reference & Large-scale (MoE) \\
\bottomrule
\end{tabular}
\end{table*}

\subsection{Ground truth aggregation} 
\label{sec:gt_aggregation}
Table \ref{tab:annotated_examples} illustrates three examples of how ground truth is obtained with different agreement types. Row 1 illustrates complete consensus across all three annotators. Row 2 demonstrates pairwise agreement; consensus labels across annotators 1–2 (democrat, outgroup, villain) and 1–3 (democrat, outgroup, loser) are merged via intersectional agreement into democrat, outgroup, villain/loser. Row 3 shows an instance of zero cross-annotator consensus; no ground truth can be established, so the meme is discarded.

\begin{table*}[!t]
\centering
\small
\setlength{\tabcolsep}{4pt}
\renewcommand{\tabularxcolumn}[1]{m{#1}}

\caption{Meme annotated examples showing agreement types and ground truth aggregation.}
\label{tab:annotated_examples}

\begin{tabularx}{\textwidth}{>{\centering\arraybackslash}m{4.6cm} @{\hspace{0.8cm}} X X X X}
\toprule
\textbf{Meme example} & \textbf{Annotator 1} & \textbf{Annotator 2} & \textbf{Annotator 3} & \textbf{Ground Truth} \\
\midrule

% Row 1
{\includegraphics[width=4.5cm, height=4cm, keepaspectratio]{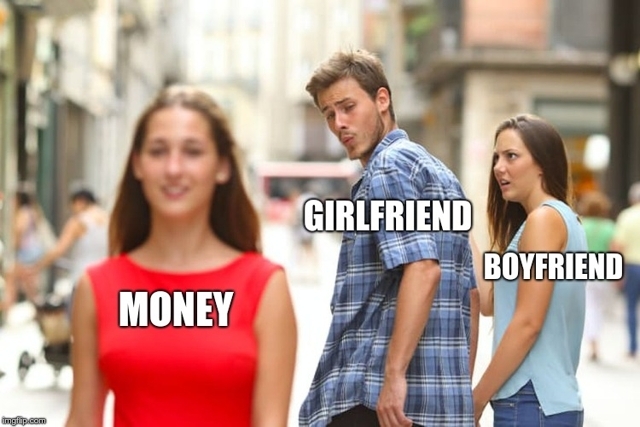}} 
& men \newline ingroup \newline victim \par\addvspace{0.4em} women \newline outgroup \newline villain 
& men \newline ingroup \newline victim \par\addvspace{0.4em} women \newline outgroup \newline villain 
& men \newline ingroup \newline victim \par\addvspace{0.4em} women \newline outgroup \newline villain 
& men \newline ingroup \newline victim \par\addvspace{0.4em} women \newline outgroup \newline villain \par\addvspace{0.4em} \textit{(perfect \newline agreement)} \\
\addlinespace[1em]

% Row 2
{\includegraphics[width=4cm, height=4cm, keepaspectratio]{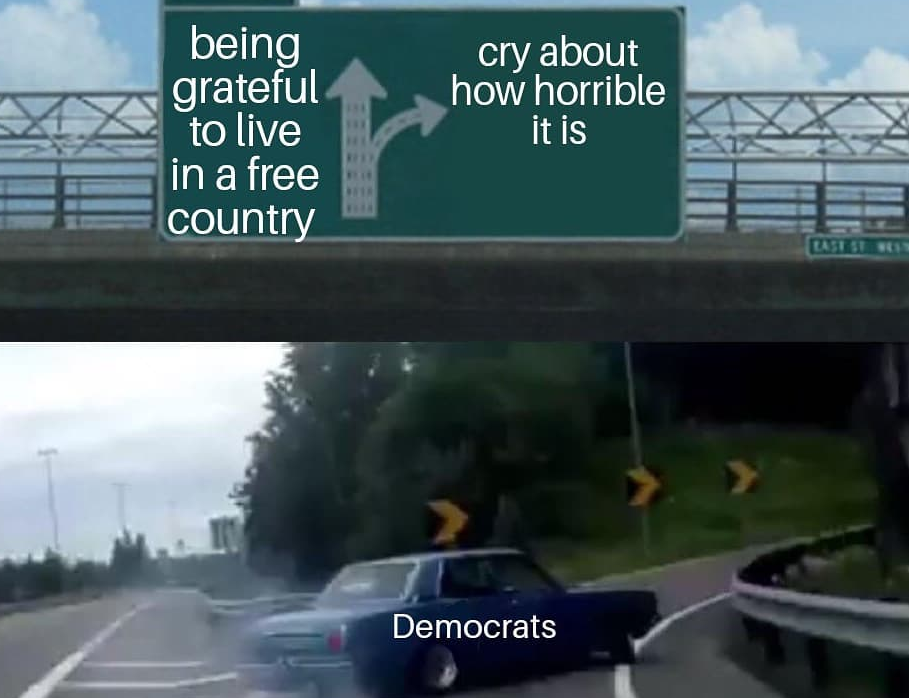}} 
& democrat \newline outgroup \newline villain/loser 
& democrat \newline outgroup \newline villain \par\addvspace{0.6em} republican \newline ingroup \newline sage 
& democrat \newline outgroup \newline loser 
& democrat \newline outgroup \newline villain/loser \par\addvspace{0.6em} \textit{(pairwise \newline agreement)} \\
\addlinespace[1em]

% Row 3
{\includegraphics[width=4cm, height=4cm, keepaspectratio]{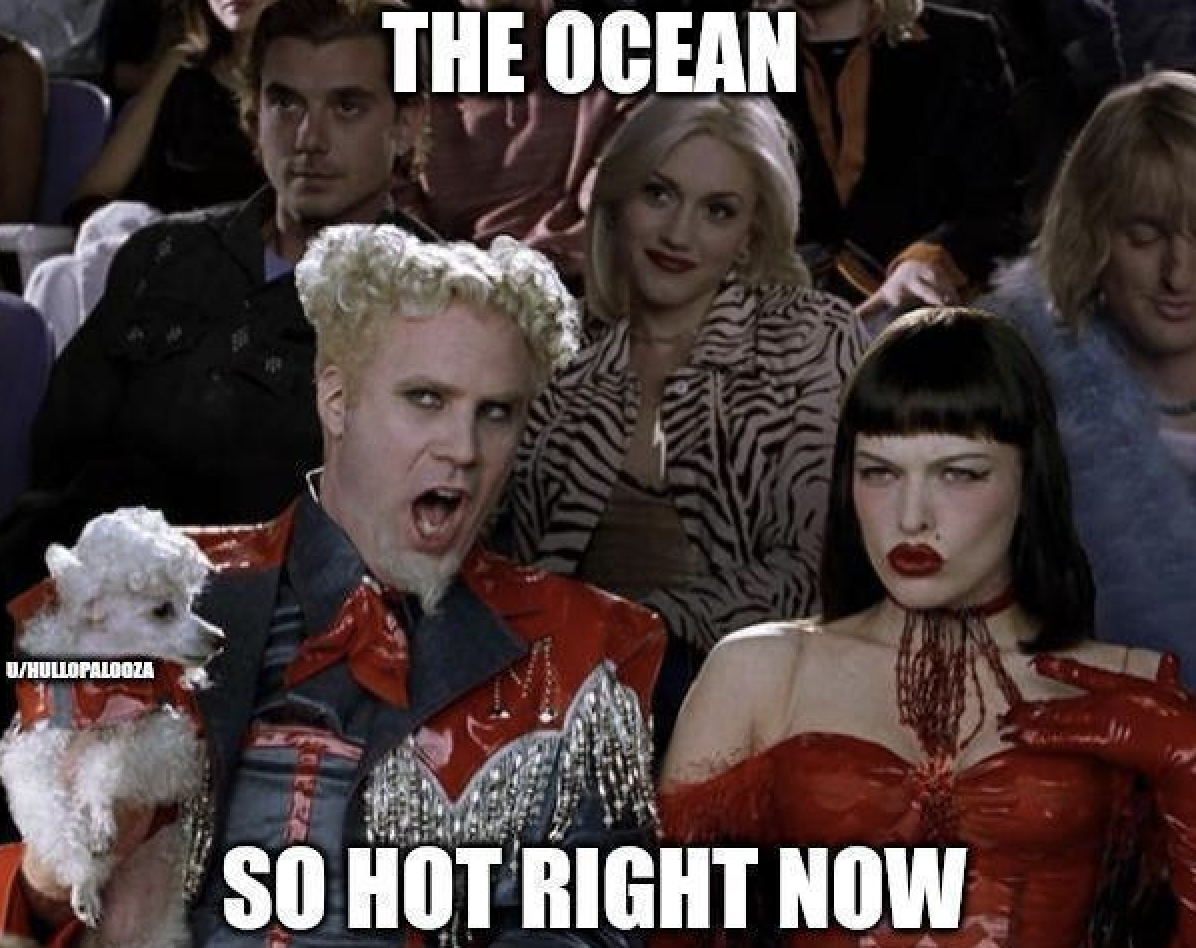}} 
& climate activist \newline ingroup \newline sage 
& climate denier \newline outgroup \newline fool 
& climate activist \newline ingroup \newline winner 
& no ground truth \par\addvspace{0.6em} \textit{(no agreement)} \\

\bottomrule
\end{tabularx}
\end{table*}

\section{Aditional results}

\subsection{Label frequency}
\label{sec:label_frequency}

\begin{figure*}[ht]
\centering
\makebox[\textwidth][c]{\includegraphics[width=0.9\textwidth]{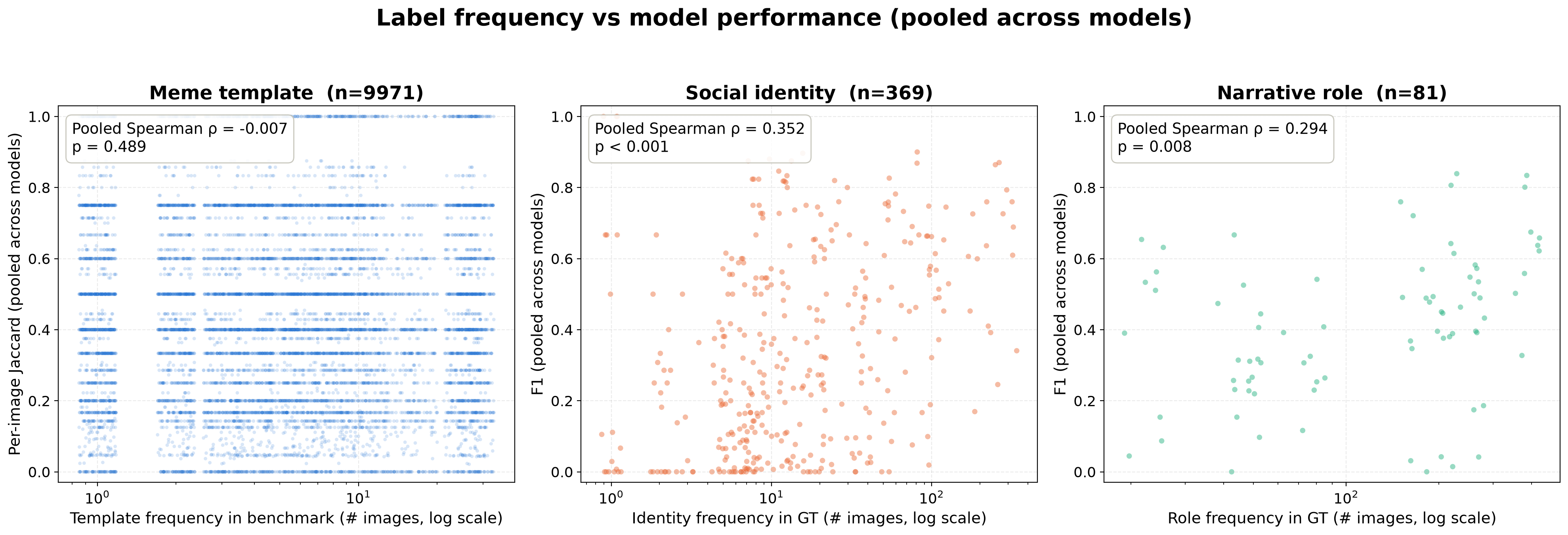}}
\caption{Label frequency and model performance across evaluation dimensions}.
\label{fig:frequency_performance}
\end{figure*}

Scatter plots in Figure \ref{fig:frequency_performance} show the relationship between ground-truth label frequency and model performance, pooled across evaluated models. Each point represents a model–label pair for Social Identity and Narrative Role, and a model–image pair for Meme Template. Frequencies are shown on a logarithmic scale. Performance is measured using F1 for Social Identity and Narrative Role and per-image Jaccard similarity for Meme Template. Spearman's $\rho$ and corresponding $p$-values are reported for each dimension.

\subsection{Error analysis}
\label{sec:error_analysis_appendix}

\begin{figure*}[tp]
\centering

\refstepcounter{subsection}
\subsection*{\thesubsection\texorpdfstring\quad {Error Analysis}}
\label{sec:error_analysis}
%\addcontentsline{toc}{subsection}{\thesubsection\texorpdfstring\quad Error Analysis}
\centering
\makebox[\textwidth][c]{\includegraphics[width=\textwidth]{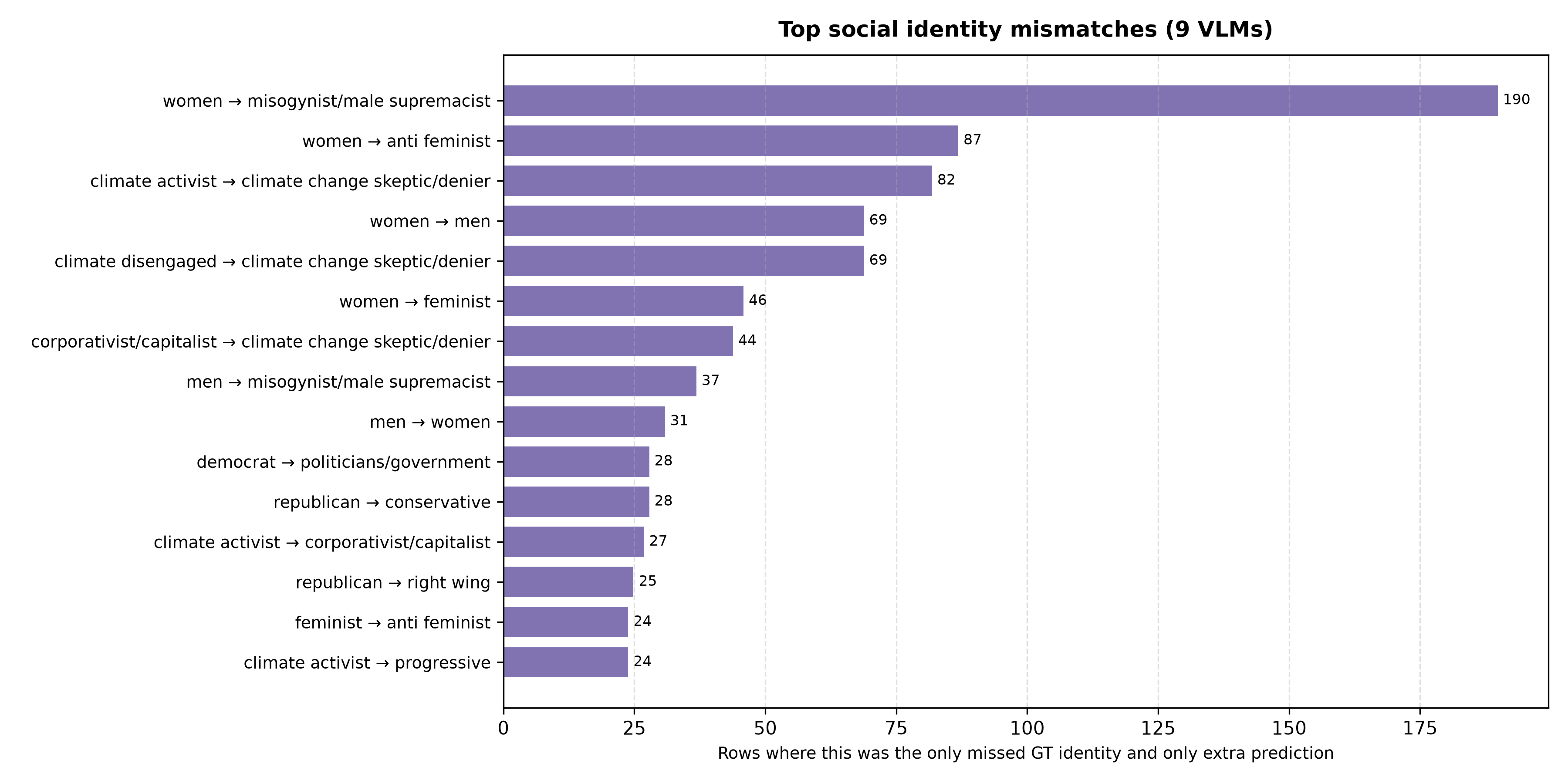}}
\caption{Top social identity mismatches accross 9 VLMs.}
\label{fig:social_identity_mismatches}
\end{figure*}

\paragraph{Social identity mismatches} Figure \ref{fig:social_identity_mismatches} shows the top social identity mismatches across the 9 VLMs. The frequency of cases in which models fail to identify a single ground-truth identity and instead predict one incorrect identity is notable. VLMs frequently demonstrate extreme antagonist substitutions, such as predicting "misogynist/male supremacist" (190 cases) or "anti feminist" (87 cases) for "women," and "climate change skeptic/denier" (82 cases) for "climate activist." Related domain-level substitutions are also common, including predictions of "conservative" (28 cases) or "right wing" (25 cases) instead of "republican." Social identity errors for VLMs involve confusing a target identity with a closely related identity within the same domain, and appear to generally be more susceptible to hallucinations than to omissions.

\begin{figure*}[ht]
\centering
\makebox[\textwidth][c]{\includegraphics[width=\textwidth]{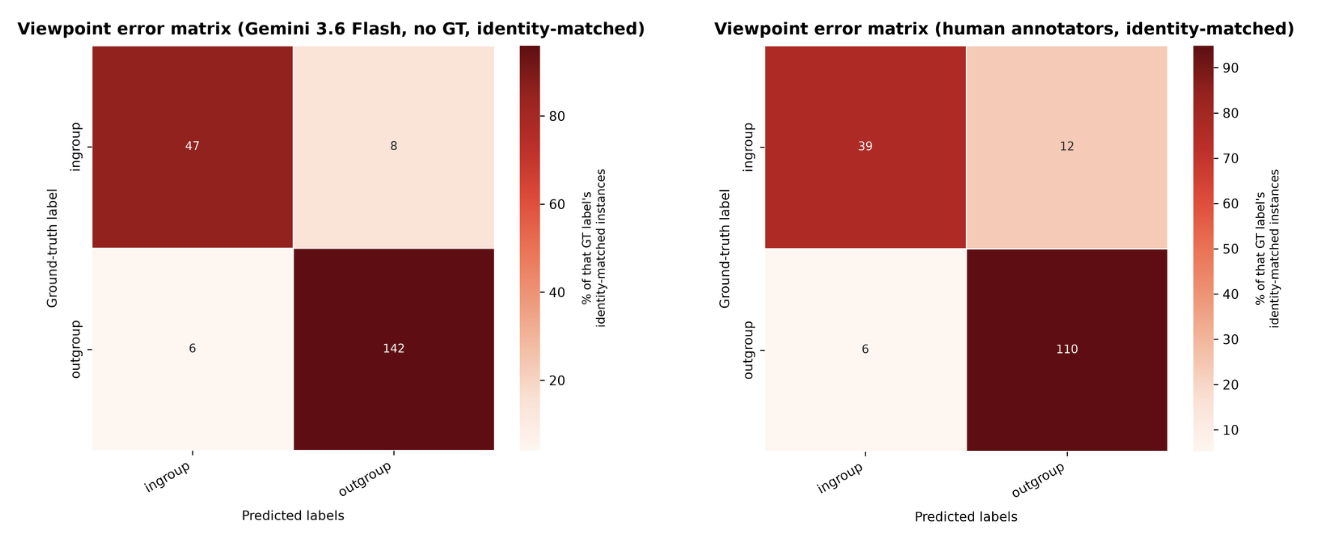}}
\caption{Viewpoint error matrices for VLM classification.}
\label{fig:viewpoint_error_matrix}
\end{figure*}

\paragraph{Viewpoint errors} Figure \ref{fig:viewpoint_error_matrix} presents the viewpoint error matrices for Gemini 3.6 Flash and human evaluators. The analysis indicates that Gemini demonstrates a slight bias toward outgroup classification, misclassifying ingroups as outgroups in 15\% of cases (8 instances out of 55 total ground-truth ingroups), compared to 4\% in the reverse direction (6 instances out of 148 total ground-truth outgroups). Human annotators exhibit a similar asymmetry, misclassifying ingroups as outgroups in 24\% of cases (12 instances out of 51 total ground-truth ingroups) and outgroups as ingroups in 5\% of cases (6 instances out of 116 total ground-truth outgroups). Considering the approximately 2.3–2.7:1 outgroup-to-ingroup imbalance in the dataset, these results suggest that both the model and human evaluators tend to default to the majority outgroup class, consistent with the findings of RQ3.

\begin{figure*}[t]
\centering
\makebox[\textwidth][c]{\includegraphics[width=\textwidth]{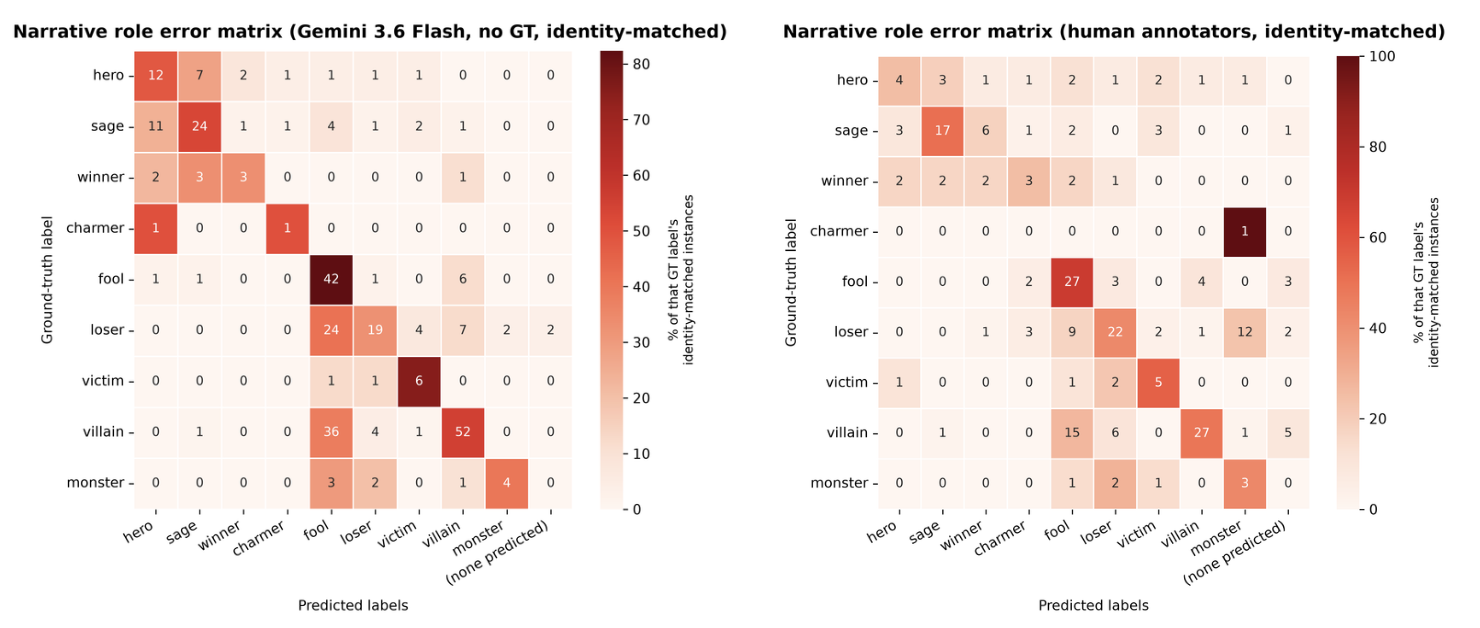}}
\caption{Narrative role error matrices for VLM classification.}
\label{fig:narrative_roles_error_matrix}
\end{figure*}

\paragraph{Narrative role errors} Figure \ref{fig:narrative_roles_error_matrix} presents the narrative role error matrices for Gemini 3.6 Flash and human evaluators. The narrative role error analysis for Gemini 3.6 Flash and human evaluators indicates a mismatch between two clusters: negatively connotated identities (villain, fool, loser, victim, monster) are predominantly confused with one another, and a similar pattern is observed among positively connotated identities (hero, sage, winner, charmer). For Gemini, the label fool functions as the primary catch-all for negative identities, frequently attracting misclassifications from ground truth (GT) villain (38%, 36 instances out of 94 GT villains) and GT loser (40%, 24 instances out of 60 GT losers). Additionally, GT sage is misclassified as hero in 25% of cases (11 instances out of 44 GT sages), compared to 55% correct sage predictions (24 instances). Human annotators exhibit a comparable valence-level pattern, but their error distribution is more diffuse, with a greater tendency to leave predictions unassigned (none predicted). In both Gemini and human evaluations, errors are largely restricted to positive or negative sub-clusters, suggesting that both models and annotators consistently recognize broader moral framing but encounter difficulty with finer-grained narrative role distinctions.

\begin{table*}[t]
\centering
\small
\renewcommand{\arraystretch}{0.85}
\begin{threeparttable}
\caption{Comparison of gemini-3.6-flash and human annotators in the setting without ground-truth context, evaluated on the same subset of 214 memes.}
\label{tab:gemini_vs_human}
\begin{tabularx}{\linewidth}{@{} l *{6}{Y} @{}}
\toprule
\multirow{2}{*}{\textbf{Model}} & \multicolumn{3}{c}{\textbf{Micro Metrics}} & \multicolumn{3}{c}{\textbf{Macro Metrics}} \\
\cmidrule(lr){2-4} \cmidrule(lr){5-7}
 & {\textbf{Precision}} & {\textbf{Recall}} & {\textbf{F1}} & {\textbf{Precision}} & {\textbf{Recall}} & {\textbf{F1}} \\
\midrule
\multicolumn{7}{l}{\textbf{Social Identities}} \\
\texttt{gemini-3.6-flash} & \bfseries 51.69 & \bfseries 77.73 & \bfseries 62.09 & \bfseries 42.98 & \bfseries 63.12 & \bfseries 46.15 \\
Human annotators & 49.85 & 64.84 & 56.37 & 34.40 & 43.60 & 35.86 \\
\midrule
\multicolumn{7}{l}{\textbf{Viewpoints}} \\
\texttt{gemini-3.6-flash} & 78.09 & \bfseries 87.35 & \bfseries 82.46 & 74.81 & \bfseries 83.41 & \bfseries 78.87 \\
Human annotators & \bfseries 81.17 & 76.68 & 78.86 & \bfseries 77.64 & 73.11 & 75.30 \\
\midrule
\multicolumn{7}{l}{\textbf{Narrative Roles}} \\
\texttt{gemini-3.6-flash} & \bfseries 46.25 & \bfseries 72.35 & \bfseries 56.43 & \bfseries 43.89 & \bfseries 71.64 & \bfseries 50.99 \\
Human annotators & 44.78 & 50.38 & 47.42 & 36.64 & 42.77 & 36.97 \\
\midrule
\multicolumn{7}{l}{\textbf{Averages}\tnote{$\dagger$}} \\
\texttt{gemini-3.6-flash} & \bfseries 58.68 & \bfseries 79.14 & \bfseries 66.99 & \bfseries 53.89 & \bfseries 72.72 & \bfseries 58.67 \\
Human annotators & 58.60 & 63.97 & 60.88 & 49.56 & 53.16 & 49.38 \\
\bottomrule
\end{tabularx}
\begin{tablenotes}
\small
\item[$\dagger$] Average metrics are calculated as the arithmetic mean across Social Identity, Viewpoint, and Narrative Roles tasks.
\end{tablenotes}
\end{threeparttable}
\end{table*}

\subsection{Ablation study: template knowledge as information intervention}
\label{sec:ablation_study}

Table \ref{tab:template_info_examples} presents three examples of narrative role classification, both with and without template information. These examples illustrate potential factors contributing to the observed decline in model performance when template information is included. In each instance, the model accurately predicts the roles without template information but fails when template descriptions are provided. In the first example, the template specifies that one character is slapped, which likely prompts the model to assign roles associated with inferiority and submission, specifically fool and loser. In the second example, the template suggests the meme pertains to either high or low intelligence; the model interprets this as low intelligence and assigns the role of fool. In the third example, the description indicates that the meme's function is "giving advice," which the model appears to interpret as a sign of wisdom, and assigns the role of the sage, although the meme is actually demeaning toward the woman.

\section{Recruitment and payment} Three students were recruited as annotators through a recruitment platform and completed the hiring process at the academic institution, providing consent to the working conditions. They were employed for three months, working 8 to 12 hours per week at a rate of 20€ per hour. Each student received 2400€ at the end of the contract.

\begin{table*}[t]
\centering
\tiny
\renewcommand{\arraystretch}{1.15}
\caption{Qualitative comparison of model predictions with (\textbf{w/}) and without (\textbf{w/o}) template descriptions in the prompt.}
\label{tab:template_info_examples}
\begin{tabular}{m{3.0cm} m{2.0cm} m{3.0cm} m{1.5cm} >{\centering\arraybackslash}m{0.7cm} >{\centering\arraybackslash}m{1.2cm} >{\centering\arraybackslash}m{1.2cm}}
\toprule
& & & & & \multicolumn{2}{c}{\textbf{Predictions}} \\
\cmidrule{6-7}
\textbf{Image} & \textbf{Template} & \textbf{Template Description} & \textbf{Identity} & \textbf{GT} & \textbf{w/o Template} & \textbf{w/ Template} \\
\midrule
% --- Row 1 ---
\vspace{0.1cm}\includegraphics[width=\linewidth]{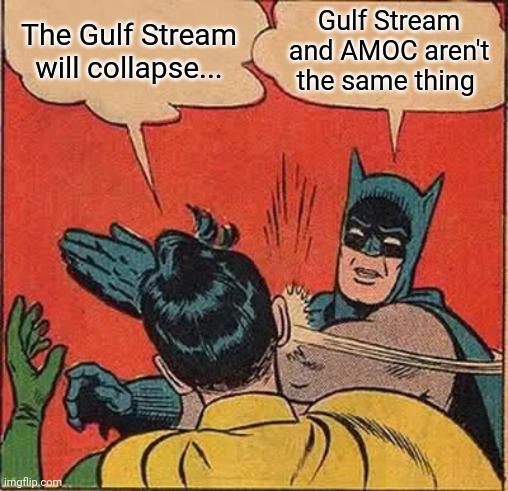}\vspace{0.1cm} &
Batman Slapping Robin &
Batman Slapping Robin, also known as ``My Parents Are Dead,'' is an exploitable image of the DC Comics superhero Batman slapping his protégé Robin in mid-conversation. Taken from a comic book published in 1965, the single-panel illustration has inspired a series of parodies featuring custom-captioned speech bubbles on 4chan, Reddit and elsewhere. &
climate activist &
sage &
sage &
fool, loser \\
\midrule
% --- Row 2 ---
\vspace{0.1cm}\includegraphics[width=\linewidth]{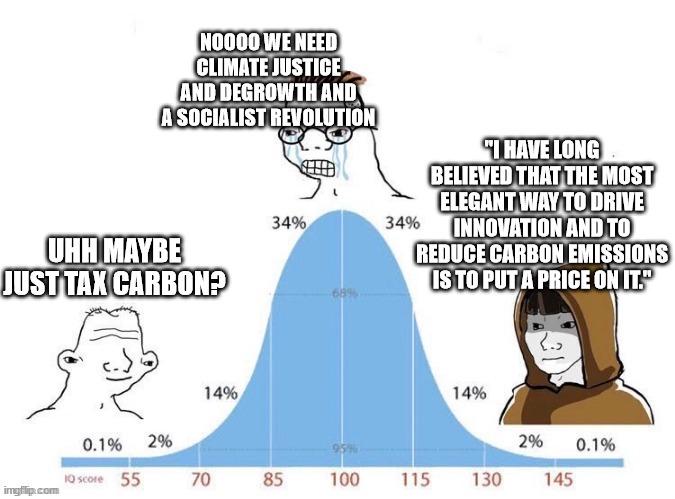}\vspace{0.1cm} &
IQ Bell Curve / Midwit &
IQ Bell Curve, also known as Midwit or simply Bell Curve Meme, refers to a series of memes that use the IQ distribution diagrams to mock one of the three main groups represented by the diagram, usually the largest group of people towards the center of the curve who possess average intelligence. The concept of IQ Bell Curve memes utilizes the horseshoe theory, implying that the groups with low intelligence and high intelligence often choose to follow the same goals while being guided by different reasoning. Those who possess average intelligence are referred to as ``midwits.'' While there were many versions of memes illustrating the concept, a variant of the meme using three different Wojaks gained particular popularity in the 2020s on X / Twitter and Reddit. &
climate activist &
sage &
sage &
fool \\
\midrule
% --- Row 3 ---
\vspace{0.1cm}\includegraphics[width=\linewidth]{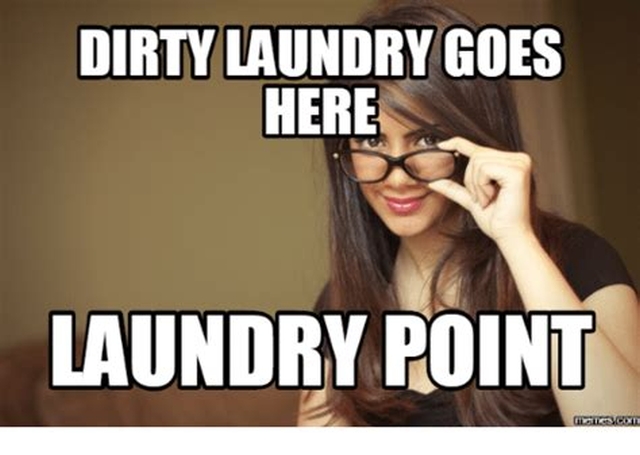}\vspace{0.1cm} &
Actual Sexual Advice Girl &
Actual Sexual Advice Girl (ASAG) is an advice animal image macro series featuring a photograph of a brunette woman wearing a pair of black-rimmed glasses accompanied by captions relaying sexual advice to both men and women. &
women &
loser &
loser &
sage \\
\bottomrule
\end{tabular}
\end{table*}

\clearpage

\begin{table*}[t]
\centering
\small % Increased font size for better legibility
\renewcommand{\arraystretch}{0.85}
\begin{threeparttable}
\caption{Performance evaluation of vision-language models grouped by category.}
\label{tab:experiment_results}
\begin{tabularx}{\linewidth}{@{} l *{6}{Y} @{}}
\toprule
\multirow{2}{*}{\textbf{Model}} & \multicolumn{3}{c}{\textbf{Micro Metrics}} & \multicolumn{3}{c}{\textbf{Macro Metrics}} \\
\cmidrule(lr){2-4} \cmidrule(lr){5-7}
 & {\textbf{Precision}} & {\textbf{Recall}} & {\textbf{F1}} & {\textbf{Precision}} & {\textbf{Recall}} & {\textbf{F1}} \\
\midrule
\multicolumn{7}{l}{\textbf{Social Identities}} \\
\texttt{deepseek-vl2} & 11.65 & 33.23 & 17.26 & 11.25 & 39.72 & 13.07 \\
\texttt{gemini-3.6-flash} & 55.79 & 76.31 & 64.46 & 45.58 & 73.55 & 51.99 \\
\texttt{glm-4.1v-9b-thinking} & 57.02 & 45.97 & 50.90 & 45.28 & 42.81 & 38.60 \\
\texttt{kimi-k3} & 55.19 & \bfseries 77.50 & \bfseries 64.47 & 44.99 & \bfseries 77.60 & \bfseries 52.78 \\
\texttt{molmo-7b-d-0924} & 3.94 & 36.34 & 7.11 & 4.16 & 27.99 & 5.80 \\
\texttt{pixtral-12b-2409} & 47.09 & 58.02 & 51.99 & 36.83 & 47.33 & 35.58 \\
\texttt{qwen2.5-vl-7b-instruct} & 27.49 & 30.30 & 28.82 & 40.58 & 37.88 & 26.41 \\
\texttt{qwen3-vl-8b-instruct} & 44.37 & 55.37 & 49.26 & 39.66 & 50.46 & 36.35 \\
\texttt{qwen3-vl-8b-thinking} & \bfseries 59.74 & 51.07 & 55.07 & \bfseries 47.99 & 47.33 & 43.42 \\
\midrule
\multicolumn{7}{l}{\textbf{Viewpoints}} \\
\texttt{deepseek-vl2} & 69.62 & 92.94 & 79.61 & 67.18 & 90.89 & 77.10 \\
\texttt{gemini-3.6-flash} & \bfseries 94.13 & \bfseries 93.70 & \bfseries 93.91 & \bfseries 93.55 & \bfseries 92.86 & \bfseries 93.20 \\
\texttt{glm-4.1v-9b-thinking} & 83.45 & 80.79 & 82.10 & 81.45 & 77.93 & 79.61 \\
\texttt{kimi-k3} & 93.46 & 93.32 & 93.39 & 92.60 & 92.62 & 92.61 \\
\texttt{molmo-7b-d-0924} & 60.65 & 92.33 & 73.21 & 60.01 & 91.25 & 71.08 \\
\texttt{pixtral-12b-2409} & 79.73 & 80.64 & 80.18 & 77.29 & 78.72 & 77.98 \\
\texttt{qwen2.5-vl-7b-instruct} & 81.81 & 75.53 & 78.55 & 79.00 & 69.18 & 72.78 \\
\texttt{qwen3-vl-8b-instruct} & 77.04 & 78.97 & 77.99 & 76.11 & 80.56 & 77.05 \\
\texttt{qwen3-vl-8b-thinking} & 82.67 & 76.77 & 79.61 & 80.51 & 75.63 & 77.93 \\
\midrule
\multicolumn{7}{l}{\textbf{Narrative Roles}} \\
\texttt{deepseek-vl2} & 34.92 & 52.60 & 41.98 & 47.00 & 41.90 & 34.44 \\
\texttt{gemini-3.6-flash} & \bfseries 59.98 & 76.58 & \bfseries 67.27 & \bfseries 59.82 & 75.75 & \bfseries 64.14 \\
\texttt{glm-4.1v-9b-thinking} & 41.66 & 63.63 & 50.35 & 41.20 & 60.49 & 46.55 \\
\texttt{kimi-k3} & 50.44 & \bfseries 82.42 & 62.58 & 48.11 & \bfseries 81.01 & 57.88 \\
\texttt{molmo-7b-d-0924} & 20.09 & 16.80 & 18.29 & 17.31 & 15.49 & 11.70 \\
\texttt{pixtral-12b-2409} & 39.17 & 49.10 & 43.58 & 41.65 & 46.81 & 40.60 \\
\texttt{qwen2.5-vl-7b-instruct} & 32.96 & 39.74 & 36.03 & 34.86 & 32.86 & 23.52 \\
\texttt{qwen3-vl-8b-instruct} & 35.68 & 65.55 & 46.21 & 41.30 & 55.53 & 41.74 \\
\texttt{qwen3-vl-8b-thinking} & 40.38 & 59.72 & 48.18 & 41.54 & 58.57 & 45.51 \\
\midrule
\multicolumn{7}{l}{\textbf{Averages}\tnote{$\dagger$}} \\
\texttt{deepseek-vl2} & 38.73 & 59.59 & 46.28 & 41.81 & 57.50 & 41.54 \\
\texttt{gemini-3.6-flash} & \bfseries 69.97 & 82.20 & \bfseries 75.21 & \bfseries 66.32 & 80.72 & \bfseries 69.78 \\
\texttt{glm-4.1v-9b-thinking} & 60.71 & 63.46 & 61.12 & 55.98 & 60.41 & 54.92 \\
\texttt{kimi-k3} & 66.36 & \bfseries 84.41 & 73.48 & 61.90 & \bfseries 83.74 & 67.76 \\
\texttt{molmo-7b-d-0924} & 28.23 & 48.49 & 32.87 & 27.16 & 44.91 & 29.53 \\
\texttt{pixtral-12b-2409} & 55.33 & 62.59 & 58.58 & 51.92 & 57.62 & 51.39 \\
\texttt{qwen2.5-vl-7b-instruct} & 47.42 & 48.52 & 47.80 & 51.48 & 46.64 & 40.90 \\
\texttt{qwen3-vl-8b-instruct} & 52.36 & 66.63 & 57.82 & 52.36 & 62.18 & 51.71 \\
\texttt{qwen3-vl-8b-thinking} & 60.93 & 62.52 & 60.95 & 56.68 & 60.51 & 55.62 \\
\bottomrule
\end{tabularx}
\begin{tablenotes}
\small
\item[$\dagger$] Average metrics are calculated as the arithmetic mean across Social Identity, Viewpoint, and Narrative Roles tasks.
\end{tablenotes}
\end{threeparttable}
\end{table*}

\end{document}